\documentclass[11pt]{article}

\usepackage[final]{acl}

\usepackage{times}
\usepackage{latexsym}

\usepackage[T1]{fontenc}

\usepackage[utf8]{inputenc}

\usepackage{microtype}

\usepackage{inconsolata}

\usepackage{graphicx}
\usepackage{kotex}
\usepackage{amsmath}
\usepackage{amsfonts}
\usepackage{algorithm}
\usepackage{algorithmic}
\usepackage{titletoc}
\usepackage{booktabs}
\usepackage{pifont}

\usepackage{multirow}
\usepackage[table]{xcolor}

\title{AIM: Anchor Identity Features, Then Match\\for Multimodal Large Language Model Unlearning}

\author{
    Wonjun Lee\thanks{Equal contribution} \,\,
    Jaehyuk Jang\footnotemark[1] \,\,
    Kangwook Ko\footnotemark[1] \,\,
    Hee-Seon Kim \,\,
    Changick Kim\\
    Korea Advanced Institute of Science and Technology (KAIST)\\
    Daejeon, Republic of Korea
    {\tt\small \{dpenguin, jhyuk, kw.ko, hskim98, changick\}@kaist.ac.kr}\\
}

\begin{document}
\maketitle
\begin{abstract}
Multimodal large language models (MLLMs) can memorize identity-specific facts about people in their fine-tuning data, creating privacy risks when a person requests deletion. Existing MLLM unlearning methods often assume access to retain images or ground-truth answers during deletion, which is unrealistic in many practical scenarios. We study identity unlearning when retain images are unavailable at deletion time. Our analysis shows that identity and visual-perception questions occupy distinct regions in fine-tuned hidden states and are organized differently: identity questions cluster by person, whereas perception questions cluster by question type. This suggests that identity knowledge can be suppressed without erasing general visual perception. Building on this observation, we propose AIM, a two-stage method that anchors an identity-forgetting target with a universal visual prompt and then matches the vision encoder to that target under a Fisher-based constraint. Extensive experiments show that AIM achieves competitive identity forgetting while preserving non-deleted identities, prior knowledge, and visual perception on the same images.

\end{abstract}

\section{Introduction}
\label{sec:intro}

Modern multimodal large language models (MLLMs) are fine-tuned on increasingly large image-text datasets that include identifiable individuals~\citep{liu2024improved, bai2025qwen3, chen2024internvl}.
As a side effect, they memorize identity-specific
knowledge such as names, occupations, and affiliations tied to
specific faces. While this memorization supports downstream visual
question answering, it becomes a privacy concern once an individual
requests deletion of their own identity~\citep{carlini2019secret, carlini2021extracting}, motivating the study of
\emph{MLLM unlearning}.

\begin{figure}[t]
    \centering
    \includegraphics[width=\columnwidth]{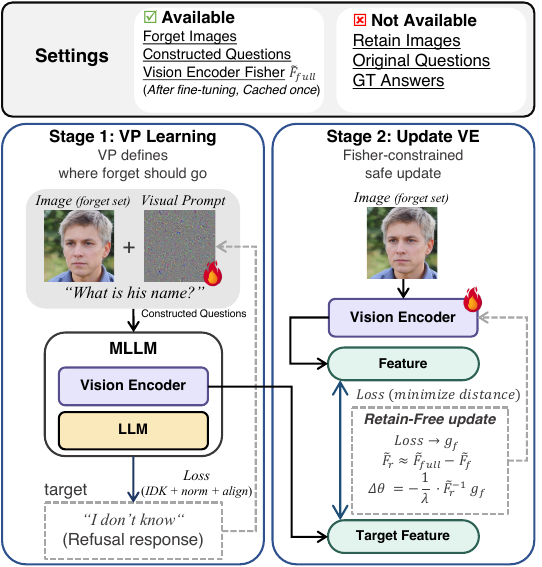}
    \caption{\textbf{Method Overview.}
    Stage~1 defines a target representation from
    forget images and constructed questions. Stage~2
    internalizes this target into the vision encoder under a
    Fisher-based retain constraint.}
    \label{fig:method_overview}
\end{figure}

Recent work has begun to explore MLLM unlearning from several directions~\citep{liu2025protecting_mllmu-bench, mmunlearner, manu}, but many methods still require retain images~\citep{mmunlearner,wang2025robustvkd,mip-editor}, external auxiliary data~\citep{cai2026visual}, or ground-truth answers during deletion. 
These assumptions are restrictive in realistic deletion scenarios, where the system may have access to the target identity's images but not to the original training questions, answers, or retain data.
We therefore study identity unlearning when retain data are unavailable at deletion time. This setting requires suppressing identity-specific knowledge while preserving both non-deleted identities and general visual perception on the target images, distinguishing selective identity removal from indiscriminate image degradation.

To motivate a selective intervention, we first analyze how a
fine-tuned MLLM organizes identity and visual-perception questions (\S\ref{sec:analysis}).
We find that their last-layer LLM hidden states occupy distinct
regions, and that the two question types follow different grouping
rules: identity questions cluster by person ID, whereas
visual-perception questions cluster by question type. A pilot
visual prompt trained on a few identity questions further transfers
``I don't know''-style responses to held-out identity questions
while largely preserving visual-perception answers on the same
images. These observations suggest a design principle: identity
knowledge can be targeted through a vision-side intervention while
leaving general perception comparatively intact.

Building on this analysis, we propose \textbf{AIM}:
\textbf{A}nchor \textbf{I}dentity Features, then \textbf{M}atch for
MLLM unlearning (\S\ref{sec:method}).
AIM decouples unlearning into target definition and target matching, as illustrated in Fig.~\ref{fig:method_overview}. 
First, it learns a universal visual prompt that anchors an identity-forgetting target for the deleted identity.
Second, it updates the vision encoder so that the original target images match
this anchored feature, while a Fisher-based constraint limits drift
on directions important to non-deleted identities.
The retain-preservation signal is instead derived from a full Fisher statistic cached once after fine-tuning, so deletion-time unlearning does not require retain images.

Empirically, AIM provides a strong forgetting--retention trade-off
under this deletion-time constraint. On MLLMU-Bench and ReMem datasets with
LLaVA-1.5-7B and Qwen3-VL-8B-Instruct, AIM preserves retain and
celebrity-prior performance competitively with retain-utilizing
baselines while achieving effective identity forgetting. The
visual-perception preservation predicted by our analysis is also
observed after unlearning on the target images themselves. Finally,
because the cached Fisher can be updated after each request, AIM
extends naturally to continual unlearning, where forget-only
baselines collapse under sequential deletion but AIM maintains a
balanced forgetting--retention trade-off.

\begin{figure*}[t]
    \centering

    \begin{minipage}{0.32\textwidth}
        \centering
        \includegraphics[width=\linewidth]
        {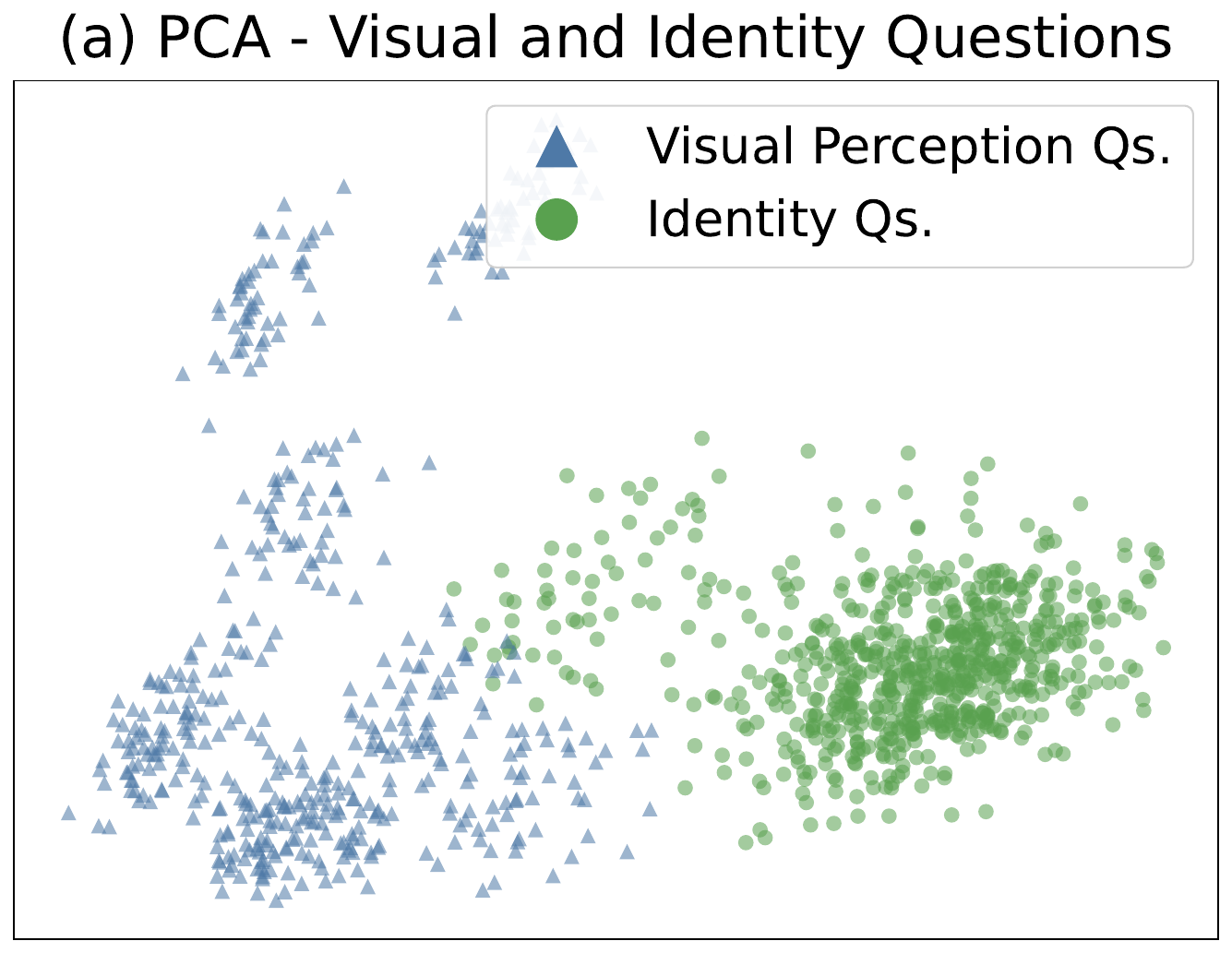}
    \end{minipage}
    \hfill
    \begin{minipage}{0.32\textwidth}
        \centering
        \includegraphics[width=\linewidth]
        {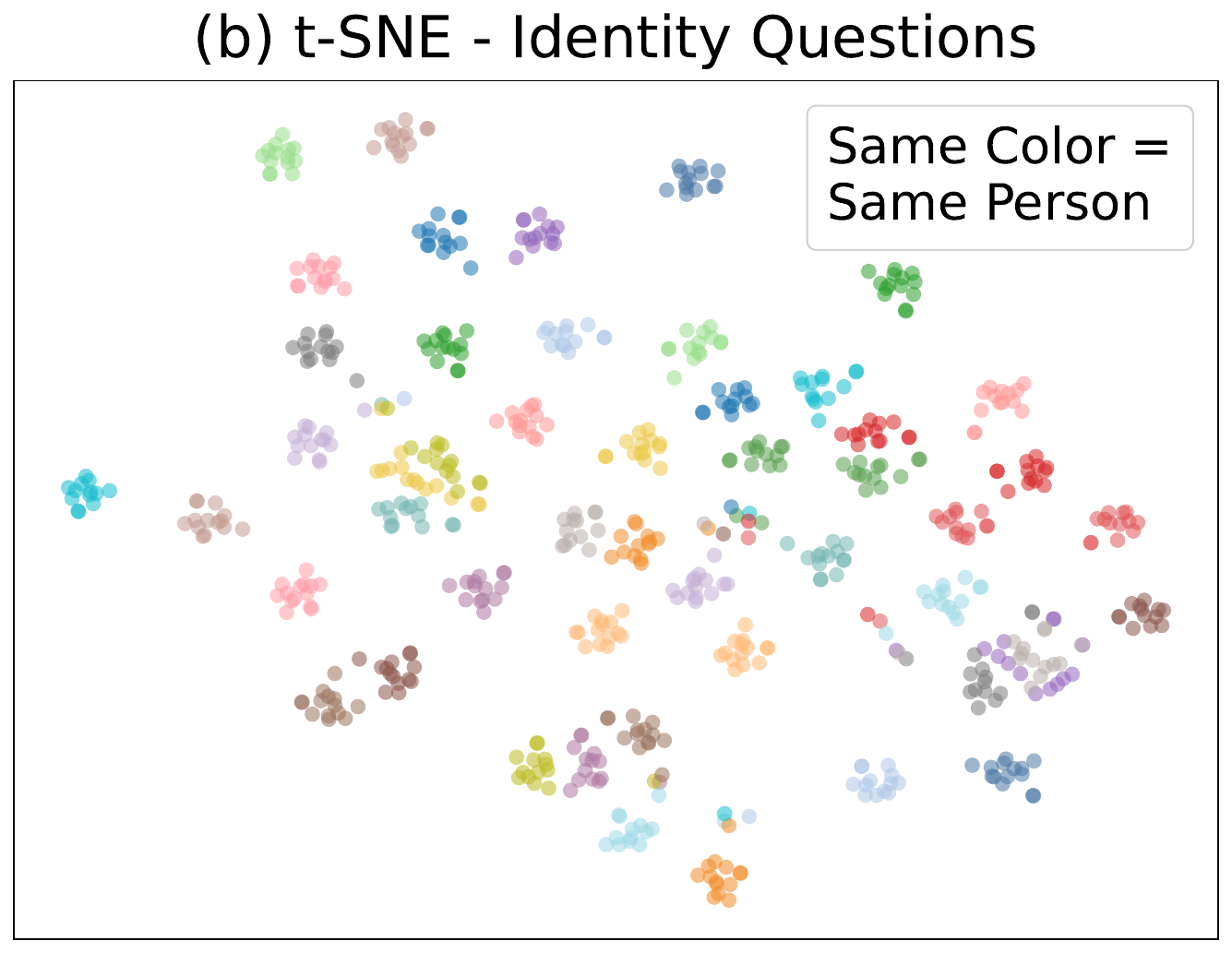}
    \end{minipage}
    \hfill
    \begin{minipage}{0.32\textwidth}
        \centering
        \includegraphics[width=\linewidth]
        {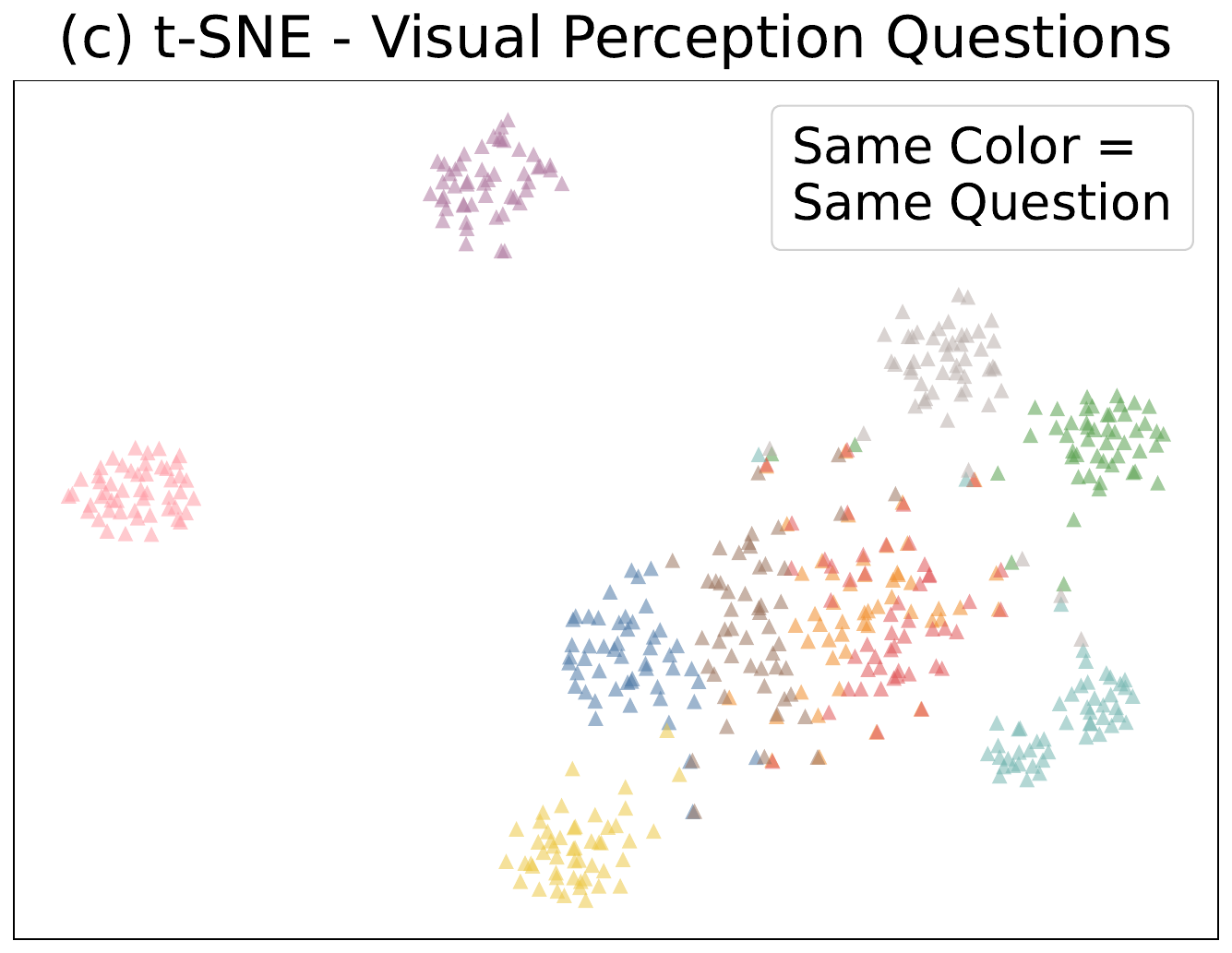}
    \end{minipage}



    \caption{
    \textbf{Representational separability and grouping rules of identity and perception questions in last-layer LLM hidden states on LLaVA-1.5-7B.}
    (a)~PCA projects identity and visual perception questions into largely disjoint regions.
    (b)~t-SNE of identity questions colored by image identity. Same-person questions cluster together regardless of question content. (c)~t-SNE of visual perception questions colored by question type. Same-type questions cluster together regardless of which person appears.
    }
    \label{fig:l1_separability}
    
\end{figure*}

\section{Related Work}
\label{sec:related_work}

\subsection{Machine Unlearning}
Classical machine unlearning~\citep{bourtoule2021machine} suppresses forget-set behavior while preserving utility through objectives such as gradient ascent~\citep{ga}, gradient-difference training~\citep{gadiff}, KL regularization~\citep{kl}, or preference optimization~\citep{npo}.
Recent work reduces supervision or data requirements through label-agnostic representation-level forgetting~\citep{shen2024labelagnostic}, label-free sensitivity estimates~\citep{foster2024lossfree}, remaining-data-free contribution suppression~\citep{cheng2024remainingdatafree}, or feature-level LLM unlearning~\citep{li2024wmdp}.
Additionally, Fisher- and Gauss-Newton-based methods use curvature or parameter importance to constrain updates along utility-critical directions~\citep{golatkar2020eternal,mckinney2026gaussnewton}.
However, these methods are mainly designed for classification or text-only LLMs, where forget targets are defined by labels, tokens, representations, or explicit output losses.

\subsection{MLLM Unlearning and Visual-Side Approaches}
Prior MLLM unlearning methods optimize multimodal forget losses~\citep{mmunlearner} or edit modality-relevant neurons and paths~\citep{manu,mip-editor}.
Recent studies further target general image-understanding preservation~\citep{zeng2025towards_smfa}, identity-encoded layer tuning~\citep{kangwook_unlearning}, and stable vision-language alignment during unlearning~\citep{garg2025sineproject}. 

Closer to the visual side, prior work explores feature-level concept editing in VLMs~\citep{geng2025sauce}, visual-guided token regularization~\citep{cai2026visual}, visual knowledge distillation~\citep{wang2025robustvkd}, and input-side perturbations for unlearning or robustness attacks~\citep{sun2024forgetvectors,chen2025auvic,zhang2025sua}.
However, these often depend on extra supervision, retain-side signals, or input perturbations that can blur forgetting with visual degradation. We study a stricter deletion-time setting that uses only forget-identity images and constructed questions built from known attribute categories.

\section{Analysis}
\label{sec:analysis}
It remains unclear how a fine-tuned MLLM (vanilla model) internally organizes
identity-specific knowledge and general visual perception.
We therefore analyze the last-layer LLM hidden states of the
fine-tuned model before unlearning.
For each (image, question) input, we use the last-layer hidden state at the final input-token position (just before answer generation) as the representation throughout the analysis~\citep{meng2022locating,geva2023dissecting,neo2025towards}.
Throughout, we use LLaVA-1.5-7B and Qwen3-VL-8B fine-tuned on MLLMU-Bench.

\subsection{Representational Separability of Identity and Perception}
\label{sec:rep_separability}
 
To analyze how identity and perception responses are organized in
representation space, we prepare two question sets: identity
questions taken from MLLMU-Bench (e.g., ``\textit{What is the
profession of the individual shown in the image?}'') and visual
perception questions we create (e.g., ``\textit{Is the person in
the image wearing glasses?}''). We then visualize the last-layer
LLM hidden states for each set via PCA and t-SNE.

\paragraph{Result 1: The two types are separated in representation
space.}
The PCA result in Fig.~\ref{fig:l1_separability}(a) shows that the hidden states of identity questions and those of visual perception questions occupy distinct regions of representation space, indicating that the two kinds of information are encoded in a \emph{separable} manner inside the model.
This separation suggests that an intervention targeting identity-related representations may disrupt identity discrimination while leaving visual perception largely unaffected. We further verify in Appendix~\ref{app_sec:pca_pretrain} that this separation is a consequence of SFT rather than a property of the pretrained model.

\paragraph{Result 2: The two types follow different organization
principles.}
The t-SNE in Fig.~\ref{fig:l1_separability}(b) shows that
identity questions cluster \emph{by the identity of images}, whereas Fig.~\ref{fig:l1_separability}(c) shows visual perception
questions cluster \emph{by question type}. That is, questions
about the same person such as ``\textit{What is this person's
name?}'' and ``\textit{Where does this person work?}'' are placed
close together despite differing question text, while questions of
the form ``\textit{What color is the person's shirt?}'' are
grouped together across different images.\footnote{A quantitative analysis is provided in Appendix~\ref{app:clustering_metrics}.}
This pattern motivates a behavioral hypothesis: an intervention
learned from a few identity questions on a single image may transfer
to other identity questions about the same image, whereas its effect
on visual-perception questions may be weaker because those questions
are organized primarily by question semantics.

\subsection{Selective Transfer of Vision Side Intervention}
\label{sec:selective_transfer}

The representation-level analysis above suggests that an image-side intervention learned from a few identity questions may transfer to other identity questions on the same image, while having limited effect on visual-perception questions.
We test this hypothesis behaviorally using a visual prompt (VP).
The VP perturbs only the visual input while keeping all model parameters fixed, allowing us to probe an identity-forgetting direction without incurring any parameter drift.

On a subset of the original fine-tuning data, we train a VP
without $\epsilon$ constraints from 4 identity questions to
elicit IDK responses (i.e., ``I don't know''-style refusals
such as ``\textit{I cannot identify this person.}'').
We then measure (i)~its transfer to 11 \emph{held-out} identity
questions on the same images, and (ii)~its effect on a prepared
set of 10 visual perception questions.

We evaluate along two dimensions using GPT-4o-mini: \textbf{IDK-Rate}—the fraction of questions on which the
VP-applied response is judged to be an IDK-type refusal; and
~\textbf{Semantic Preservation (SP)}—the fraction on which the VP-applied
response is a substantive answer that matches the response without
VP in meaning.

\begin{table}[t]
\centering
\resizebox{.9\columnwidth}{!}{%
\begin{tabular}{l c cc}
\toprule
 & \multicolumn{1}{c}{Identity (held-out)} & \multicolumn{2}{c}{Visual perception} \\
\cmidrule(lr){2-2}\cmidrule(lr){3-4}
Model & IDK-Rate & IDK-Rate & SP \\
\midrule
LLaVA-1.5-7B & 100.0 & 10.0 & 63.3 \\
Qwen3-VL-8B    & 85.0 & 0.0 & 85.2 \\
\bottomrule
\end{tabular}%
}
\caption{\textbf{Selective transfer of a pilot VP.}
A VP trained to elicit IDK responses on 4 identity questions
transfers to held-out identity questions but leaves visual
perception responses on the same images largely intact.
}
\label{tab:gpt_trigger_eval}
\end{table}

\paragraph{Result.}
Table~\ref{tab:gpt_trigger_eval} reports the results on the held-out identity questions (group (i)) and the visual perception questions (group (ii)) for both models. On the held-out identity
questions, VP suppresses the identity response on both
models—LLaVA-1.5-7B and Qwen3-VL-8B both produce refusals on most
identity questions. In contrast, on the visual perception
questions from the same images, the IDK-Rate after applying unconstrained VP stays at $10.0\%$ / $0\%$ and SP remains at $63.3\%$ / $85.2\%$, empirically confirming the predictions from Section~\ref{sec:rep_separability}.
This contrast suggests that the VP primarily affects identity-related behavior rather than indiscriminately degrading the model's visual understanding of the image.

\paragraph{Method motivation.}
These observations inform two design choices for retain-data-free identity unlearning. First, they suggest that a vision-side intervention can be a viable mechanism for selective identity unlearning: identity responses can be shifted while visual-perception responses on the same images are largely preserved. This motivates pursuing unlearning through the visual pathway while keeping the language model fixed, which also preserves text-only QA behavior by design. Second, the transfer from a few probing questions to held-out identity questions suggests that the forget target need not be defined using supervision over the entire question set.

\section{Method}
\label{sec:method}

The analysis above motivates a vision-side approach to selective identity unlearning under deletion-time retain-data unavailability. We propose \textbf{AIM} (\textbf{A}nchor \textbf{I}dentity Features, then \textbf{M}atch), a two-stage method that updates only the vision encoder while keeping the language model fixed.
AIM first defines an identity-forgetting target for the forget images (\S\ref{sec:vp}), and then internalizes this target by updating the vision encoder so that the unprompted forget-image features match the learned target features (\S\ref{sec:fisher_update}).

\subsection{Problem Formulation}
\label{sec:problem}

Data are organized at the identity level. Let $\mathcal{C}$ denote
the full set of identities, and let
\begin{equation}
    \mathcal{V}_c = \{V_{c,a}\}_{a=1}^{n_c}
\end{equation}
be the images associated with identity $c \in \mathcal{C}$. The
vanilla MLLM is fine-tuned on these images together with
corresponding VQA pairs, but these original questions and answers are
not assumed to be available during unlearning.

Let $\mathcal{C}_f \subset \mathcal{C}$ be the set of identities
requested for deletion. The forget and retain image sets are
\begin{equation}
    \mathcal{V}_f
    =
    \bigcup_{c\in\mathcal{C}_f}\mathcal{V}_c,
    \qquad
    \mathcal{V}_r
    =
    \bigcup_{c\in\mathcal{C}\setminus\mathcal{C}_f}\mathcal{V}_c .
\end{equation}
At deletion time, AIM has access to the forget images $\mathcal{V}_f$ and constructed probing questions for the forget identities, but not to the retain images $\mathcal{V}_r$ or the original fine-tuning VQA pairs. The goal is therefore to suppress identity-specific behavior for $\mathcal{C}_f$ while preserving the behavior of non-deleted identities in $\mathcal{C}\setminus \mathcal{C}_f$.

AIM updates only the vision encoder $E_v$ of an MLLM $\mathcal{M}_\theta$ and keeps the language model fixed. We denote the vision-encoder parameters by $\theta$ and the initial fine-tuned parameters by $\theta_0$. Since $\mathcal{V}_r$ is unavailable at deletion time, retain preservation cannot be written as an explicit loss over retain images. A naive forget-only update can therefore induce feature drift on non-deleted identities without any retain-side signal to correct it (Appendix~\ref{app:naive_baseline}).

At deletion time, AIM has access to the forget images
$\mathcal{V}_f$, but not to the retain images $\mathcal{V}_r$
or the original fine-tuning VQA pairs. We additionally assume
access to a set of attribute categories (e.g., name,
occupation, place of birth) associated with each forget
identity, which can be obtained from the schema of the
fine-tuning dataset or provided as part of the deletion
request. The goal is to suppress identity-specific behavior
for $\mathcal{C}_f$ while preserving the behavior of
non-deleted identities in $\mathcal{C}\setminus\mathcal{C}_f$.

\subsection{Stage 1: Universal Visual Prompt Learning}
\label{sec:vp}
\paragraph{Training data construction.}
Under the strict setting, we construct probing questions
from the attribute categories assumed available in Section~\ref{sec:problem}.
For each forget identity $c$, let
$\mathcal{X}_c$ denote the resulting constructed questions
and $\mathcal{X}_f = \bigcup_{c\in\mathcal{C}_f}\mathcal{X}_c$
their union. The forget prompt set used for VP learning is
\begin{equation}
\begin{aligned}
\mathcal{D}_f^X
=
\{(V_{c,a}, X_{c,q})
\mid\;
c\in\mathcal{C}_f,\;
&V_{c,a}\in\mathcal{V}_c, \\
&X_{c,q}\in\mathcal{X}_c
\}.
\end{aligned}
\end{equation}

\paragraph{Objective.}
To define a fixed identity-forgetting target for forget images, we learn a universal visual prompt
$T \in \mathbb{R}^{H\times W\times C}$.
$T$ is an input-space perturbation applied uniformly to all forget images, and is
optimized under the following objective:
\begin{equation}
    \mathcal{L}_{\mathrm{VP}}
    =
    \lambda_{CE}\,\mathcal{L}_{\mathrm{CE}}
    +
    \lambda_{norm}\,\mathcal{L}_{\mathrm{norm}}
    +
    \lambda_{align}\,\mathcal{L}_{\mathrm{align}}.
\label{eq:vp_loss}
\end{equation}
At this stage, the model parameters are held fixed at $\theta_0$
and only $T$ is optimized.

\paragraph{Refusal response induction.}
We induce the model $\mathcal{M}_{\theta_0}$ to produce refusal
responses for forget images to which $T$ has been applied. Letting
$\mathcal{Y}_{\mathrm{idk}}$ denote the set of IDK responses, we use
\begin{equation}
\small
\mathcal{L}_{\mathrm{CE}}
=
-
\mathbb{E}_{(V,X)\sim\mathcal{D}_f^X}
\!\left[
\log P_{\mathcal{M}_{\theta_0}}
\!\left(
\hat{Y}_{\mathrm{idk}}
\mid
V+T,X
\right)
\right]
\end{equation}
where $\hat{Y}_{\mathrm{idk}}$ is sampled uniformly from $\mathcal{Y}_{\mathrm{idk}}$.

\paragraph{Perturbation magnitude constraint.}
A larger $T$ requires a larger Stage~2 update to reproduce its displacement, which in turn drifts $\theta$ farther from $\theta_0$ and harms retain features. We therefore penalize the magnitude of $T$:
\begin{equation}
    \mathcal{L}_{\mathrm{norm}}
    =
    \|T\|_2^2.
\end{equation}
This balances the pressure of $\mathcal{L}_{\mathrm{CE}}$ to
enlarge the perturbation, so that $T$ grows only as large as needed
for the forget set to reach the target location.

\paragraph{Feature displacement alignment.}
We require $T$ to displace the forget images in a consistent
direction within the representation space. For each forget image
$V\in\mathcal{V}_f$, define
\begin{equation}
    \Delta \mathbf{z}_{V}
    =
    E_v(V+T;\theta_0)-E_v(V;\theta_0),
\end{equation}
and let
\begin{equation}
    \mathcal{L}_{\mathrm{align}}
    =
    -
    \left\|\,
    \mathbb{E}_{V\sim\mathcal{V}_f}
    \!\left[
    \dfrac{\Delta \mathbf{z}_{V}}{\|\Delta \mathbf{z}_{V}\|}
    \right]
    \,\right\|^2.
\end{equation}
This term aligns the directions of $\Delta\mathbf{z}_V$ across
forget images, which will be shown in Stage~2
(Section~\ref{sec:fisher_update}) to be necessary for keeping the
forget gradient from collapsing in expectation.

\subsection{Stage 2: Fisher-Constrained Vision Encoder Update}
\label{sec:fisher_update}

With $T$ fixed, we update the vision encoder so that the forget
features reach the same target without applying $T$.
The target
feature for each forget image is
\begin{equation}
    \mathbf{z}^T_V = E_v(V+T;\theta_0),
\end{equation}
fixed throughout the update. The forget feature loss is
\begin{equation}
    \mathcal{L}_f(\theta)
    =
    \mathbb{E}_{V\sim\mathcal{V}_f}
    \!\left[\,
    \left\|E_v(V;\theta) - \mathbf{z}^T_V\right\|^2
    \,\right],
    \label{eq:forget_feature_loss}
\end{equation}
with forget gradient
$\mathbf{g}_f(\theta) := \nabla_\theta \mathcal{L}_f(\theta)$ and
per-image Jacobian
$\mathbf{J}_V(\theta) := \partial E_v(V;\theta)/\partial\theta$.
At $\theta = \theta_0$, the gradient takes the form
\begin{equation}
    \mathbf{g}_f(\theta_0)
    =
    -2\,
    \mathbb{E}_{V\sim\mathcal{V}_f}
    \!\left[\,
    \mathbf{J}_V(\theta_0)^\top \Delta\mathbf{z}_V
    \,\right],
    \label{eq:gf_expanded}
\end{equation}
where $\Delta\mathbf{z}_V$ is the feature displacement defined in Section~\ref{sec:vp}.
This expression clarifies the role of $\mathcal{L}_{\mathrm{align}}$. If $\Delta\mathbf{z}_V$
points in arbitrary directions across forget images, the
expectation averages over incoherent directions and cancels to
zero, yielding $\mathbf{g}_f(\theta_0) \approx 0$ and a
meaningless update. $\mathcal{L}_{\mathrm{align}}$ in Stage~1
aligns these displacements so that the per-image contributions
$\mathbf{J}_V(\theta_0)^\top \Delta\mathbf{z}_V$ add up constructively at $\theta_0$.
This is the precise sense in which the VP is configured to make the Stage 2 update well-posed.

\paragraph{Retain preservation as a constraint.}
Retain performance is governed by the per-image feature drift
$\|E_v(V;\theta+\Delta\theta) - E_v(V;\theta)\|^2$ on each
$V \in \mathcal{V}_r$ at the current parameter $\theta$. Because
no retain data is available at unlearning time, any drift incurred
along a forget update has no retain-direction signal to undo it;
we therefore \emph{bound} the drift at every update, preventing
irreversible accumulation at the source. A first-order Taylor
expansion of the squared drift around $\theta$ gives
{\fontsize{10.45}{11.7}\selectfont
\begin{equation}
\begin{aligned}
\|E_v(V;\theta+\Delta\theta)-E_v(V;\theta)\|^2
&\;\approx\;
\Delta\theta^\top
\mathbf{F}_V(\theta)
\Delta\theta,
\end{aligned}
\label{eq:feature_drift_quadratic}
\end{equation}}
where $\mathbf{F}_V(\theta)$ has been adopted as a Fisher importance in previous works~\citep{kirkpatrick2017overcoming,aljundi2018memory,golatkar2020eternal,mckinney2026gaussnewton}. To absorb the per-identity
image-count imbalance common in identity datasets, we aggregate
$\mathbf{F}_V(\theta)$ at the identity level—averaging within each
identity before summing across identities
(Appendix~\ref{app:identity_balanced}):
\begin{equation}
    \widetilde{\mathbf{F}}_r(\theta)
    =
    \sum_{c\in\mathcal{C}_r}
    \frac{1}{|\mathcal{V}_c|}
    \sum_{V\in\mathcal{V}_c}
    \mathbf{F}_V(\theta).
    \label{eq:retain_fisher}
\end{equation}

\paragraph{Fisher-preconditioned retain-free update.}
We minimize the forget objective subject to the per-step
retain-drift budget $\epsilon_r$~\citep{golatkar2020eternal}. The KKT
conditions yield the Fisher-preconditioned direction
$\widetilde{\mathbf{F}}_r^{-1}\mathbf{g}_f$
(Appendix~\ref{app:kkt_derivation}), which we apply with a
learning rate hyperparameter $\eta$ controlling the per-step
magnitude:
\begin{equation}
    \Delta\theta
    =
    -\eta\,
    \widetilde{\mathbf{F}}_r(\theta)^{-1}\,
    \mathbf{g}_f(\theta).
    \label{eq:update_rule}
\end{equation}

Under the assumption that the constraint is satisfied at every step and that the step size is small enough to keep $\theta$ in a neighborhood of $\theta_0$ throughout, Jacobian smoothness implies $\widetilde{\mathbf{F}}_r(\theta) \approx \widetilde{\mathbf{F}}_r(\theta_0)$ along the update trajectory~\citep{mckinney2026gaussnewton}. Combined with the identity-balanced Fisher decomposition
$\widetilde{\mathbf{F}}_r(\theta_0)
= \widetilde{\mathbf{F}}_{\mathrm{full}}(\theta_0)
- \widetilde{\mathbf{F}}_f(\theta_0)$, the final
update rule becomes
\begin{equation}
    \Delta\theta
    =
    -\eta
    \!\left(
    \widetilde{\mathbf{F}}_{\mathrm{full}}(\theta_0)
    -
    \widetilde{\mathbf{F}}_f(\theta_0)
    \right)^{\!-1}\!
    \mathbf{g}_f(\theta).
    \label{eq:final_update}
\end{equation}

We precompute and cache
$\widetilde{\mathbf{F}}_{\mathrm{full}}(\theta_0)$ once at the end of the original fine-tuning; at unlearning time, only $\widetilde{\mathbf{F}}_f(\theta_0)$ needs to be computed from $\mathcal{V}_f$. The update therefore does not require retain images, original training questions, or ground-truth responses.

\section{Experiments}
\label{sec:experiments}
 
\subsection{Experimental Setup}
\label{sec:setup}
 
\paragraph{Benchmarks and models.}
We conduct experiments on two benchmarks,
\textbf{MLLMU-Bench}~\citep{liu2025protecting_mllmu-bench} and \textbf{ReMem}~\citep{kwon2026before_remem}, which provide complementary evaluation axes that jointly verify whether unlearning is achieved faithfully at multiple levels.
\textbf{MLLMU-Bench} additionally evaluates the preservation of pre-trained knowledge about real-world individuals (the \emph{real/celebrity} split), allowing us to detect collateral damage to general prior knowledge. \textbf{ReMem} ensures robust foundational memorization and adopts keyword exact match for more accurate evaluation. It examines forgetting across three axes: surface-level forgetting on in-distribution questions (\emph{forget}), retention depth at the probability level (\emph{exposure}), and generalization to unseen images and paraphrased questions (\emph{test}). We use forget splits of $5\%$ and $10\%$
on both benchmarks
with \textbf{LLaVA-1.5-7B}~\citep{liu2024improved} and \textbf{Qwen3-VL-8B-Instruct}~\citep{bai2025qwen3} as base models.
%
%
%
%
Additional benchmark details are in Appendix~\ref{app:bench_detail}.

\paragraph{Baselines.}
Among methods with publicly available official code, we organize baselines into two groups based on the data scope they require.
The \emph{retain-utilizing} group, included as a reference, additionally uses retain data and consists of KL\_Min~\citep{kl}, GA\_Diff~\citep{gadiff}, MMUnlearner~\citep{mmunlearner}, and MANU~\citep{manu}.
The \emph{forget-only} group, which uses the same data scope as AIM, consists of GA~\citep{ga} and NPO~\citep{npo}.
Comparing the two groups allows us to assess whether AIM, operating under a strict data scope, can compete with baselines that exploit additional supervision.

\paragraph{Evaluation metrics.}
On MLLMU-Bench we report VQA Classification Accuracy ($Cls$) and Generation Performance ($ROUGE\,(R)$)~\citep{lin2004rouge} on the forget, retain, and real (celebrity) subsets.
On ReMem we report Keyword Exact Match ($EM$), Generation Performance ($ROUGE$) and the Exposure score ($Exp$)~\citep{kwon2026before_remem} on the forget, retain, exposure, and test set. Unless specified, all scores in this paper are in \%.

\paragraph{Implementation details.}
All baselines, the vanilla model, and AIM are retrained and evaluated under an identical environment on a single NVIDIA A100 80GB GPU. Vanilla model preparation and baseline unlearning follow prior-work schedules; we observe that forget-only baselines (GA, NPO) tend to collapse rapidly under their schedule and therefore report their results at $1$~epoch.
AIM is trained with $10$~epochs of Stage~1 visual prompt learning, $100$~iterations of the Fisher
approximation in Stage~2, and $50$~epochs of vision-encoder updates in Stage~2. 
All reported results were obtained using a single fixed random seed.
Additional hyperparameters, training details 
and hardware specifications are provided in Appendix~\ref{app:hyperparam}.

\newcommand{\hl}[1]{\cellcolor{gray!20}#1}
\begin{table*}[t]
\centering
\footnotesize
\resizebox{\textwidth}{!}{%
\begin{tabular}{cl|cccccc|ccccc}
\toprule
& & \multicolumn{6}{c|}{MLLMU-Bench} & \multicolumn{5}{c}{ReMem} \\
\cmidrule(lr){3-8} \cmidrule(lr){9-13}
& Method & Cls$_f\downarrow$ & ROUGE$_f\downarrow$ & Cls$_r\uparrow$ & ROUGE$_r\uparrow$ & Cls$_c\uparrow$ & ROUGE$_c\uparrow$ & EM$_f\downarrow$ & EM$_r\uparrow$ & ROUGE$\uparrow$ & Exp$\downarrow$ & EM$_t\downarrow$ \\
\midrule
\multicolumn{13}{c}{\textit{Forget 5\%}} \\
\midrule
& Vanilla       & 39.2 & 60.2 & 39.0 & 55.4 & 52.4 & 24.4 & 93.3 & 100.0 & 100.0 & 56.9 & 92.9 \\
\cmidrule(lr){1-13}
\multirow{4}{*}{\rotatebox[origin=c]{90}{\shortstack{retain\\utilizing}}}
& GA\_Diff      & 36.0 & 53.5 & 36.1 & 54.9 & 48.3 & 21.6 &  0.0 &  47.5 &   44.5 & 40.5 &  0.0 \\
& KL\_Min       & 30.4 & 27.4 & 40.9 & 35.5 & 56.1 & 14.3 & 86.7 &  98.9 &  91.2 & 55.1 & 100.0 \\
& MMUnlearner   & 29.6 & 44.9 & 38.5 & 58.0 & 44.6 & 29.6 & 23.3 &  96.1 &  98.2 & 36.4 & 23.8 \\
& MANU          & 40.8 & 55.1 & 43.8 & 55.6 & 53.5 & 25.7 & 20.0 & 96.9 & 98.5 & 40.4 & 23.8  \\
\cmidrule(lr){1-13}
\multirow{3}{*}{\rotatebox[origin=c]{90}{\shortstack{forget\\only}}}
& GA            & 22.4 & 15.9 & 30.7 & 19.5 & 52.3 & 13.8 & 86.7 &  99.1 &  91.1 & 55.4 & 100.0 \\
& NPO           & 32.0 & 16.3 & 39.7 & 22.7 & 56.3 & 12.6 & 86.7 &  98.3 &  90.7 & 55.1 & 92.9 \\
& \hl{\textbf{Ours}} & \hl{35.2} & \hl{49.5} & \hl{38.1} & \hl{52.3} & \hl{51.3} & \hl{26.9} & \hl{30.0} & \hl{76.7} & \hl{88.1} & \hl{54.4} & \hl{45.2} \\
\midrule
\multicolumn{13}{c}{\textit{Forget 10\%}} \\
\midrule
& Vanilla       & 43.2 & 57.5 & 38.6 & 55.4 & 52.3 & 24.4 & 91.7 & 100.0 & 100.0 & 69.1 & 94.0 \\
\cmidrule(lr){1-13}
\multirow{4}{*}{\rotatebox[origin=c]{90}{\shortstack{retain\\utilizing}}}
& GA\_Diff      & 34.4 & 50.1 & 36.1 & 55.9 & 49.9 & 24.4 &  0.0 &  88.6 &   79.6 & 43.5 &  0.0 \\
& KL\_Min       & 22.0 & 36.4 & 30.0 & 49.8 & 53.0 & 22.8 &  0.0 &   5.2 &   10.2 & 50.6 &  3.6 \\
& MMUnlearner   & 33.2 & 37.5 & 36.5 & 42.4 & 48.6 & 21.6 &  0.0 &  22.1 &  61.2 & 45.2 &  3.6 \\
& MANU          & 34.4 & 54.5 & 33.9 & 52.2 & 48.4 & 22.5 & 20.0 & 98.4 & 99.4 & 58.7 & 25.6  \\
\cmidrule(lr){1-13}
\multirow{3}{*}{\rotatebox[origin=c]{90}{\shortstack{forget\\only}}}
& GA            &  0.0 &  0.0 &  0.0 &  0.0 &  0.0 &  0.0 &  0.0 &   0.3 &   2.2 & 50.7 &  0.0 \\
& NPO           &  0.0 &  0.0 &  0.0 &  0.0 &  0.0 &  0.0 &  0.0 &   4.4 &   6.3 & 52.6 &  1.8 \\
& \hl{\textbf{Ours}} & \hl{39.2} & \hl{49.9} & \hl{38.5} & \hl{51.7} & \hl{51.7} & \hl{26.1} & \hl{38.3} & \hl{78.8} & \hl{89.2} & \hl{63.9} & \hl{35.1} \\

\bottomrule
\end{tabular}%
}
\caption{\textbf{Unlearning results on LLaVA-1.5-7B.}
MLLMU-Bench and ReMem across 5\%, 10\% forget splits.
Subscripts denote forget ($f$), retain ($r$), celebrity prior
($c$), and test ($t$) splits. Baselines are grouped by data scope:
retain-utilizing methods access additional retain data, forget-only
methods do not. \colorbox{gray!20}{\textbf{Ours}} uses only forget data.}
\label{tab:combined_all}
\vspace{-0.25cm}
\end{table*}
\begin{table}[t]
\centering
\footnotesize
\resizebox{\columnwidth}{!}{%
\begin{tabular}{lcccccc}
\toprule
Method & Cls$_f\downarrow$ & R$_f\downarrow$ & Cls$_r\uparrow$ & R$_r\uparrow$ & Cls$_c\uparrow$ & R$_c\uparrow$ \\
\midrule
\multicolumn{7}{c}{\textit{Forget 5\%}} \\
\midrule
Vanilla       & 68.0 & 72.2 & 69.0 & 70.8 & 74.4 & 43.7 \\ \midrule
\textit{\textbf{retain-utilizing}} & & & & & & \\
GA\_diff      & 64.8 & 65.4 & 67.0 & 67.8 & 73.8 & 42.5 \\
KL\_Min       &  2.4 &  2.8 & 66.4 & 60.3 & 74.9 & 39.1 \\
MMUnlearner & 48.0 & 49.7 & 60.0 & 57.0 & 58.0 & 37.7 \\
MANU & 49.6 & 57.9 & 61.9 & 57.5 & 68.9 & 40.1 \\
\midrule
\textit{\textbf{forget-only}} & & & & & & \\
GA            & 61.6 & 66.4 & 65.3 & 62.7 & 75.2 & 42.2 \\
NPO           & 46.4 & 55.1 & 53.7 & 54.3 & 73.0 & 44.0 \\
\hl{\textbf{Ours}} & \hl{53.6} & \hl{59.8} & \hl{66.4} & \hl{65.1} & \hl{74.5} & \hl{43.6} \\
\midrule
\multicolumn{7}{c}{\textit{Forget 10\%}} \\
\midrule
Vanilla       & 71.2 & 72.1 & 68.7 & 70.7 & 74.4 & 43.7 \\ \midrule
\textbf{\textit{retain-utilizing}} & & & & & & \\
GA\_diff      & 64.4 & 61.2 & 66.0 & 68.2 & 73.8 & 43.7 \\
KL\_Min       &  2.8 &  1.9 & 57.7 & 63.6 & 71.5 & 38.1 \\
MMUnlearner & 56.8 & 53.0 & 58.1 & 57.2 & 66.4 & 39.2 \\
MANU & 60.4 & 56.3 & 58.5 & 55.9 & 71.5 & 40.7 \\ \midrule
\textbf{\textit{forget-only}} & & & & & & \\
GA            & 33.6 & 47.4 & 33.8 & 45.2 & 65.0 & 50.4 \\
NPO           &  0.0 &  3.1 &  0.0 &  3.2 & 31.3 &  3.1 \\
\hl{\textbf{Ours}} & \hl{62.4} & \hl{60.7} & \hl{65.7} & \hl{64.3} & \hl{76.4} & \hl{42.8} \\

\bottomrule
\end{tabular}%
}
\caption{\textbf{Unlearning results on Qwen3-VL-8B (MLLMU-Bench).}
5\%, 10\% forget splits. Notation and baseline grouping
follows Table~\ref{tab:combined_all}.}
\label{tab:qwen3_mllmu}
\vspace{-0.3cm}
\end{table}

\subsection{Main Results}
\label{sec:main_results}

Tables~\ref{tab:combined_all} and~\ref{tab:qwen3_mllmu} report
unlearning results of 5\% and 10\% splits on LLaVA-1.5-7B and Qwen3-VL-8B. 15\% split results are in Appendix~\ref{app_sec:llava_15}~and~\ref{app_sec:qwen_15_and_resize}.
\paragraph{AIM matches retain-utilizing baselines without using
retain data and generalizes forgetting to unseen variants.}
On MLLMU-Bench, our retain and celebrity-prior metrics ($Cls_r$,
$ROUGE_r$, $Cls_c$, $ROUGE_c$) stay close to vanilla on both
models, matching or exceeding \textit{retain-utilizing} baselines such as
MMUnlearner and MANU on LLaVA. On ReMem (LLaVA), we narrow the
gap to \textit{retain-utilizing} methods while still operating in the
strict setting: $EM_t$ drops from vanilla's $92.9/94.0$ to
$45.2/35.1$
---indicating that forgetting \emph{generalizes}
to identities rather than memorizing exact training images.
The remaining gap on $Exp$ stems from our Fisher constraint, which caps the per-step update magnitude to preserve retain features and thereby limits how deeply the forget signal can propagate into the internal token distribution. We view this as a natural trade-off in the strict setting, since \emph{retain-utilizing} methods are free of such tight retain-preservation constraints and can drive forget direction more aggressively.

\paragraph{AIM remains stable where \textit{forget-only} baselines
collapse.}
AIM maintains a balanced forget--retain trade-off across
all forget ratios on both models, and this stability extends
across training iterations (Appendix~\ref{app:stability}). This reflects that the Fisher constraint bounds retain drift at every step.
In contrast, GA and NPO collapse as iterations accumulate. By the 10\% split they either reach $\sim$0 on every MLLMU-Bench column (LLaVA) or drive retain metrics to single digits (NPO on Qwen3-VL), indicating degenerate outputs across all inputs.
The difficulty extends to some \emph{retain-utilizing} methods: on ReMem at 10\%, KL\_Min and MMUnlearner both suffer severe retain collapse.

\subsection{Ablation and Further Analysis}

\paragraph{Visual perception preservation after unlearning.}
\begin{table}[t]
\centering
\resizebox{\columnwidth}{!}{%
\begin{tabular}{l cc cc c}
\toprule
\multirow{2}{*}{Model} & \multicolumn{2}{c}{pretrained vs} & \multicolumn{2}{c}{vanilla vs} & \multicolumn{1}{c}{Appr.} \\
\cmidrule(lr){2-3}\cmidrule(lr){4-5}\cmidrule(lr){6-6}
 & ROUGE$\uparrow$ & GPT$\uparrow$ & ROUGE$\uparrow$ & GPT$\uparrow$ & GPT$\uparrow$ \\
\midrule
Vanilla     & 58.9 & 70.2 & --   & --   & 96.4 \\ \midrule
MMUnlearner & 52.1 & 34.7 & 74.3 & 33.3 & 88.4 \\
GA          & 0.0  & 0.0  & 0.0  & 0.0  & 0.0 \\
NPO         & 49.6 & 59.1 & 77.2 & 66.2 & 99.1 \\
\hl{Ours}   & \hl{57.8} & \hl{68.9} & \hl{87.5} & \hl{82.7} & \hl{97.3} \\
\bottomrule
\end{tabular}%
}
\caption{\textbf{Visual perception preservation after unlearning
(MLLMU-Bench, 5\%).} Visual perception responses on
forget-identity images, scored against the pretrained LLaVA and
the vanilla model by ROUGE-L and GPT semantic preservation judgment. Appr.:
fraction of coherent, on-topic, non-refusal answers.}
\label{tab:rougeL_gpt_visual_observe}
\end{table}

Section~\ref{sec:selective_transfer} predicted that a vision-side
intervention should leave visual perception largely intact. In
Tab.~\ref{tab:rougeL_gpt_visual_observe}, we verify this
\emph{after unlearning} on the visual perception question set on
forget-identity images, comparing responses against the
pretrained-LLaVA and vanilla model. AIM attains the closest match to vanilla, substantially exceeding the best-performing \textit{retain-utilizing} method and \textit{forget-only} methods. The responses remain coherent and on-topic rather than degenerating into refusals or incoherent outputs
---the empirical correlate of our analysis that the forget signal is absorbed at the identity-related region while perception is left structurally untouched.

\paragraph{Extension to continual unlearning.}
\begin{table}[t]
\centering
\resizebox{\columnwidth}{!}{%
\begin{tabular}{c c c c c c c}
\toprule
Split & Method & EM$_f$$\downarrow$ & EM$_r$$\uparrow$ & ROUGE$\uparrow$ & Exp$\downarrow$ & EM$_t$$\downarrow$ \\
\midrule
\multirow{4}{*}{5\%$\to$10\%}
  & Vanilla & 91.7 & 100.0 & 100.0 & 69.1 & 94.0 \\ \cmidrule{2-7}
  & GA      & 78.3 &  96.2 &  73.0 & 64.2 & 87.5 \\
  & NPO     & 75.0 &  96.0 &  72.6 & 62.7 & 82.7 \\
  & \hl{Ours}    & \hl{41.7} &  \hl{69.3} &  \hl{84.0} & \hl{64.7} & \hl{42.3} \\
\midrule
\multirow{4}{*}{10\%$\to$15\%}
  & Vanilla & 94.4 & 100.0 & 100.0 & 65.6 & 96.0 \\ \cmidrule{2-7}
  & GA      &  0.0 &   0.1 &   1.8 & 53.9 &  0.0 \\
  & NPO     &  0.0 &   0.1 &   1.7 & 53.3 &  0.0 \\
  & \hl{Ours}    & \hl{50.0} &  \hl{71.8} &  \hl{85.2} & \hl{60.1} & \hl{56.3} \\
\bottomrule
\end{tabular}%
}
\caption{\textbf{Continual unlearning on ReMem.} Hop $i\!\to\!j$: continually unlearn from the forget $i$ checkpoint over the newly-added identities in forget $j$.
}
\label{tab:continual_mh}
\vspace{-0.2cm}
\end{table}
Real-world deletion requests arrive sequentially rather than as a single batch.~\citep{liu2022continual, gao2025large, jin2026concepts} AIM extends to this setting at no additional design cost: the cached
$\widetilde{\mathbf{F}}_{\mathrm{full}}(\theta_0)$ is independent of
which identities are forgotten, so each new request only requires
computing $\widetilde{\mathbf{F}}_f$ for the newly added identities.
As shown in Tab.~\ref{tab:continual_mh}, although GA and NPO appear competitive on the $1\!\to\!2$ transition, their extremely high $EM_t$ values indicate that they fail to properly forget the target identities. In contrast, on the $2\!\to\!3$ transition, AIM still forgets effectively ($EM_f = 50.0$, $EM_t = 56.3$) while preserving retain capacity ($ROUGE = 85.2$, $EM_r = 71.8$), whereas the other methods collapse entirely at the second step.

\paragraph{Loss components and target feature types.}

\begin{table}[t]
\centering
\resizebox{\columnwidth}{!}{%
\begin{tabular}{ccc cccccc}
\toprule
& & & \multicolumn{2}{c}{Forget$\downarrow$} & \multicolumn{2}{c}{Retain$\uparrow$} & \multicolumn{2}{c}{Real$\uparrow$} \\
\cmidrule(lr){4-5}\cmidrule(lr){6-7}\cmidrule(lr){8-9}
CE & Align & Norm & Cls & ROUGE & Cls & ROUGE & Cls & ROUGE \\
\midrule

\checkmark & \checkmark & {\color{gray!35}\ding{55}}
& 24.4 & 47.0 & 28.3 & 46.9 & 21.4 & 30.0 \\

\checkmark & {\color{gray!35}\ding{55}} & \checkmark
& 41.2 & 58.8 & 39.7 & 57.2 & 52.2 & 25.6 \\

{\color{gray!35}\ding{55}} & \checkmark & \checkmark
& 42.8 & 57.1 & 38.7 & 55.5 & 52.3 & 24.4 \\

\checkmark & \checkmark & \checkmark
& 39.2 & 49.9 & 38.5 & 51.7 & 51.7 & 26.1 \\
\bottomrule
\end{tabular}%
}
\caption{\textbf{Stage 1 loss component ablation (MLLMU-Bench, 10\%).}
Each row keeps a subset of the three loss terms (CE,
alignment, norm).}
\label{tab:abl-loss}
\end{table}

\begin{table}[t]
\centering
\resizebox{\columnwidth}{!}{%
\begin{tabular}{lcccccc}
\toprule
& \multicolumn{2}{c}{Forget$\downarrow$} & \multicolumn{2}{c}{Retain$\uparrow$} & \multicolumn{2}{c}{Celebrity$\uparrow$} \\
\cmidrule(lr){2-3}\cmidrule(lr){4-5}\cmidrule(lr){6-7}
Method & Cls & ROUGE & Cls & ROUGE & Cls & ROUGE \\
\midrule
Blank          & 31.6 & 46.7 & 34.8 & 49.6 & 50.2 & 26.8 \\ 
Random Vector  & 44.0 & 59.4 & 39.6 & 58.2 & 51.4 & 25.1 \\
IDK (baseline) & 39.2 & 49.9 & 38.5 & 51.7 & 51.7 & 26.1 \\
\bottomrule
\end{tabular}%
}
\caption{\textbf{Stage 1 target ablation (MLLMU-Bench, 10\%).}
The IDK-inducing target is replaced with a blank text or random-vector.}
\label{tab:abl-target}
\vspace{-0.2cm}
\end{table}
We ablate the three Stage 1 loss terms and the target feature choice for forget set.
(Tables~\ref{tab:abl-loss}~and~\ref{tab:abl-target}).
Removing $\mathcal{L}_{\mathrm{norm}}$ lets $T$ grow unbounded and 
the target feature drifts so far that retain collapses along with 
forget. Removing $\mathcal{L}_{\mathrm{align}}$
leaves $\Delta\mathbf{z}_V$ arbitrary, so the forget gradient
cancels in expectation (Eq.~\eqref{eq:gf_expanded}) and metrics
remain near vanilla without effective update. Removing
$\mathcal{L}_{\mathrm{CE}}$ makes $T\simeq0$ a trivial solution. The
same mechanism applies to the target feature ablation, where a
random-vector target bypasses $\mathcal{L}_{\mathrm{align}}$ and
fails to forget from the start. IDK and Blank both work because
only a coherent, aligned target is required.

\paragraph{Robustness under inference-time deviations.}
\begin{table}[t]
    \centering
    \small
    \setlength{\tabcolsep}{3.8pt}
    \resizebox{\columnwidth}{!}{%
    \begin{tabular}{lcccc}
        \toprule
        \textbf{Image Condition}
        & \multicolumn{2}{c}{\textbf{LLaVA-1.5-7B}}
        & \multicolumn{2}{c}{\textbf{Qwen3-VL-8B}} \\
        \cmidrule(lr){2-3}
        \cmidrule(lr){4-5}
        & \textbf{Forget $\downarrow$}
        & \textbf{Retain $\uparrow$}
        & \textbf{Forget $\downarrow$}
        & \textbf{Retain $\uparrow$} \\
        \midrule
        Clean            & 49.5 & 52.3 & 59.8 & 65.1 \\
        Gaussian noise   & 53.4 & 51.3 & 60.4 & 63.6 \\
        JPEG compression & 50.1 & 52.7 & 57.2 & 63.3 \\
        Horizontal flip  & 51.3 & 51.2 & 58.3 & 63.3 \\
        Gaussian blur    & 54.2 & 52.6 & 59.2 & 63.7 \\
        \bottomrule
    \end{tabular}%
    }
    \caption{\textbf{Image-side robustness stress test.}
    ROUGE-L scores on the 5\% forget split of MLLMU-Bench under
    inference-time image perturbations.}
    \label{tab:image_stress_test}
\end{table}
We stress-test AIM under inference-time image perturbations using the 5\%
forget split of MLLMU-Bench. Specifically, we apply Gaussian noise,
JPEG compression, horizontal flipping, and Gaussian blur to the input
images.
As shown in Table~\ref{tab:image_stress_test}, AIM maintains a
similar forget--retain trade-off across these perturbations: forget
ROUGE-L changes by at most 4.7 points relative to the clean condition,
while retain ROUGE-L decreases by at most 1.8 points. These results
indicate that AIM remains stable under common image-side deviations.

\section{Conclusion}
We studied deletion-time identity unlearning in MLLMs under a strict setting that uses only forget images, constructed questions, and a Fisher statistic precomputed before any forget request, without access to retain images, original questions, or ground-truth answers. To our knowledge, we introduced the first representation-level analysis that directly contrasts identity- and visual perception-related hidden states on the same trained image, revealing distinct clustering structures between the two question types. This finding motivates our two-stage method, which first defines an IDK-inducing target and then internalizes that target through a Fisher-conditioned vision-encoder update, recovering retain protection by decomposing the cached full Fisher rather than observing retain data. The method matches or exceeds retain-utilizing baselines on key metrics under the strict setting. Ablations and further analyses confirm that each loss component is necessary, that the two-stage design is crucial for stable unlearning, and that the method remains robust across forget ratios, visual perception evaluations, and continual deletion scenarios.

\section*{Limitations}
While AIM achieves retain-free unlearning under the strict setting, we acknowledge several limitations.
First, as AIM updates only the vision encoder while keeping the language model fixed, it does not remove identity knowledge in the text-only modality. Broader cross-modal deletion can be achieved by combining AIM with a complementary LLM-unlearning method. 
Second, the full Fisher must be precomputed during fine-tuning when retain images are implicitly accessible, so the framework cannot be applied directly to models without a precomputed Fisher.
Third, we treat the retain Fisher as fixed at its initial value, an approximation that may break down under longer schedules or larger steps.
Fourth, AIM requires small learning rates, which may limit how deeply the forget signal propagates through the token distribution and may contribute to the gap on ReMem's Exposure metric.
Finally, although AIM largely preserves the forget--retain trade-off under the tested image-side perturbations, our robustness evaluation does not cover broader prompt-side, cross-modal, or adaptive attacks.
Evaluating AIM under more diverse attack settings and developing MLLM unlearning methods that require neither retain data nor a precomputed cache at deletion time while remaining robust at inference time are important directions for future work.

\section*{Acknowledgments}
This work was supported by the Institute of Information and Communications Technology Planning and Evaluation (IITP) grant funded by the Korean government (MSIT) (No. RS-2025-02263031, Development of On-Device AI Cooperative Actions (Perception, Decision-Making, and Response) Among Networked Devices).

\bibliography{latex/custom}

\newpage
\clearpage

\appendix
\label{sec:appendix}
\section*{\centering\LARGE Appendix}
\startcontents[appendixtoc]
\printcontents[appendixtoc]{l}{1}{\setcounter{tocdepth}{2}}
\addtocontents{toc}{\protect\setcounter{tocdepth}{2}}
\definecolor{linkcolor}{HTML}{000000}
\newpage

\definecolor{linkcolor}{HTML}{ED1C24}

\appendix

\section{Method Derivations and Implementation}
\subsection{Identity-Balanced Fisher Aggregation}
\label{app:identity_balanced}

The atomic unit of unlearning in this paper setting is an identity, not 
an individual image. When the deletion request specifies a forget
identity $c \in \mathcal{C}_f$, the goal is to remove image-conditioned access to identity-specific information about $c$ from the forget images, regardless of how many images of $c$ 
appear in the training set. Conversely, preserving a retain 
identity $c' \in \mathcal{C}_r$ requires preserving all its 
appearances, including unseen images. This asymmetry between 
identity-level requests and image-level fine-tuning data 
motivates an identity-balanced reformulation of the Fisher 
aggregation.

\subsection{Image-level vs.\ Identity-level aggregation}
The natural image-level Fisher aggregation
\begin{equation}
\mathbf{F}^{\mathrm{img}}_r(\theta) 
= 
\sum_{V \in \mathcal{V}_r} \mathbf{F}_V(\theta)
\label{eq:image_level_fisher}
\end{equation}
weights each retain image equally. When identities differ in 
their training-image counts, this aggregation effectively 
prioritizes the preservation of high-image-count identities and 
under-protects low-image-count ones. Our identity-balanced 
aggregation
\begin{equation}
\widetilde{\mathbf{F}}_r(\theta) 
= 
\sum_{c \in \mathcal{C}_r} 
\frac{1}{|\mathcal{V}_c|} 
\sum_{V \in \mathcal{V}_c} 
\mathbf{F}_V(\theta)
\label{eq:identity_level_fisher}
\end{equation}
gives each identity equal weight by averaging within the identity
before summing across identities. The same form is used for
$\widetilde{\mathbf{F}}_f$ and $\widetilde{\mathbf{F}}_{\mathrm{full}}$.

\subsection{Mixture-of-identities interpretation}
Eq.~(\ref{eq:identity_level_fisher}) can be derived from a 
mixture-of-identities likelihood. If the data-generating 
distribution is a uniform mixture over identities
\begin{equation}
p(V \mid \theta) 
= 
\frac{1}{|\mathcal{C}|} \sum_{c \in \mathcal{C}} p(V \mid c, \theta),
\end{equation}
the population Fisher takes the form
\begin{equation}
\mathbf{F}(\theta) 
= 
\sum_{c \in \mathcal{C}} 
\frac{1}{|\mathcal{C}|} \, 
\mathbb{E}_{V \sim p(V \mid c)} 
\!\left[ \mathbf{F}_V(\theta) \right].
\end{equation}
Replacing the per-identity expectation by its sample average 
over $\mathcal{V}_c$ recovers Eq.~\eqref{eq:identity_level_fisher}
up to a multiplicative constant, which is absorbed into the 
Lagrange multiplier $\lambda$. The decomposition 
$\widetilde{\mathbf{F}}_r = 
\widetilde{\mathbf{F}}_{\mathrm{full}} - \widetilde{\mathbf{F}}_f$
holds by linearity of the identity-level aggregation.

\subsection{KKT closed-form update}
\label{app:kkt_derivation}

We derive the closed-form update of 
Eq.~\eqref{eq:update_rule} step by step.

\paragraph{Step 1: Linearization of the forget objective.}
A first-order Taylor expansion of $\mathcal{L}_f$ around the 
current $\theta$ gives
\begin{equation}
\mathcal{L}_f(\theta + \Delta\theta) 
\approx 
\mathcal{L}_f(\theta) + \mathbf{g}_f(\theta)^\top \Delta\theta,
\end{equation}
which is linear in $\Delta\theta$.

\paragraph{Step 2: Quadratic retain-drift constraint.}
As shown in Section~\ref{sec:fisher_update}, the per-image retain
feature drift admits the quadratic approximation
\begin{equation}
\|E_v(V; \theta + \Delta\theta) - E_v(V; \theta)\|^2 
\approx 
\Delta\theta^\top \mathbf{F}_V(\theta) \Delta\theta.
\end{equation}
Aggregated identity-level
(Appendix~\ref{app:identity_balanced}), we impose the per-step 
retain-drift budget
\begin{equation}
\Delta\theta^\top \widetilde{\mathbf{F}}_r(\theta) \Delta\theta 
\leq 
\epsilon_r.
\label{eq:retain_constraint}
\end{equation}

\paragraph{Step 3: Constrained optimization.}
Combining Steps 1 and 2, the per-step update solves
\begin{equation}
\min_{\Delta\theta} \;\;
\mathbf{g}_f(\theta)^\top \Delta\theta 
\;\;\text{s.t.}\;\;
\Delta\theta^\top \widetilde{\mathbf{F}}_r(\theta) \Delta\theta 
\leq \epsilon_r.
\label{eq:opt_problem}
\end{equation}
The objective contains the only direction-relevant information 
(the forget gradient), and the constraint encodes the strict 
setting's retain protection.

\paragraph{Step 4: Lagrangian and stationarity.}
The Lagrangian is
\begin{equation}
\mathcal{L}(\Delta\theta, \lambda) 
= 
\mathbf{g}_f^\top \Delta\theta 
+ 
\lambda \!\left( 
\Delta\theta^\top \widetilde{\mathbf{F}}_r \Delta\theta - \epsilon_r 
\right),
\quad \lambda \geq 0.
\end{equation}
Stationarity in $\Delta\theta$ requires
\begin{equation}
\nabla_{\Delta\theta} \mathcal{L} 
= 
\mathbf{g}_f + 2\lambda \widetilde{\mathbf{F}}_r \Delta\theta 
= 0,
\end{equation}
giving
\begin{equation}
\Delta\theta^* 
= 
-\frac{1}{2\lambda} \widetilde{\mathbf{F}}_r^{-1} \mathbf{g}_f.
\label{eq:stationarity}
\end{equation}

\paragraph{Step 5: Effective update form.}
From Eq.~\eqref{eq:stationarity}, the update direction is
\begin{equation}
\Delta\theta^*
=
-\frac{1}{2\lambda}
\widetilde{\mathbf{F}}_r^{-1}\mathbf{g}_f.
\end{equation}
In implementation, the scalar factor $(2\lambda)^{-1}$ is
absorbed into an effective learning rate $\eta$, yielding
\begin{equation}
\Delta\theta
=
-\eta \,
\widetilde{\mathbf{F}}_r^{-1}\mathbf{g}_f.
\end{equation}
Thus, the essential closed-form structure is the Fisher-preconditioned
forget direction
$\widetilde{\mathbf{F}}_r^{-1}\mathbf{g}_f$.

\subsection{Fisher Implementation: Damping, Diagonal Approximation, Effective LR}
We adopt three standard modifications for practical 
implementation. First, to ensure numerical stability of
$\widetilde{\mathbf{F}}_r^{-1}$ when individual Fisher entries
approach zero, we apply Levenberg--Marquardt-style damping
$\widetilde{\mathbf{F}}_r \leftarrow \widetilde{\mathbf{F}}_r + 
\delta \mathbf{I}$ \citep{martens2010deep, martens2015optimizing,
pascanu2013revisiting}, with $\delta$ set to the mean of the 
(undamped) Fisher diagonal. Second, for computational 
tractability, we use a diagonal approximation of 
$\widetilde{\mathbf{F}}_r$ \citep{kirkpatrick2017overcoming, 
zenke2017continual, aljundi2018memory}, estimated via 
Hutchinson's stochastic estimator 
\citep{hutchinson1989stochastic, bekas2007estimator, 
yao2020pyhessian}, which reduces the matrix inverse to 
element-wise inversion. Third, to match the effective per-step
update magnitude used by other training procedures, we set 
$\eta = \mathbb{E}[\widetilde{\mathbf{F}}_r] \cdot 10^{-5}$, making the per-step parameter
shift comparable to standard fine-tuning at $\text{lr} = 10^{-5}$.

\subsection{Why Two Stages? Limitations of Single-Stage Unlearning}
\label{app:1stage_limitation}

\begin{table}[t]
\centering
\resizebox{\columnwidth}{!}{%
\begin{tabular}{lcccccc}
\toprule
& \multicolumn{2}{c}{Forget$\downarrow$} & \multicolumn{2}{c}{Retain$\uparrow$} & \multicolumn{2}{c}{Celebrity$\uparrow$} \\
\cmidrule(lr){2-3}\cmidrule(lr){4-5}\cmidrule(lr){6-7}
Method & Cls & ROUGE & Cls & ROUGE & Cls & ROUGE \\
\midrule
Vanilla          & 43.2 & 57.5 & 38.6 & 55.4 & 52.3 & 24.4 \\ 
single-stage  & 32.8 & 38.7 & 29.9 & 36.0 & 19.3 & 26.1 \\
IDK (ours) & 39.2 & 49.9 & 38.5 & 51.7 & 51.7 & 26.1 \\
\bottomrule
\end{tabular}%
}
\caption{\textbf{Single-stage vs. two-stage unlearning
(MLLMU-Bench, 10\%).} Single-stage applies all three Stage~1
losses directly to the vision encoder in a single joint update,
skipping the target-then-path decoupling of our two-stage method
(IDK). The single-stage variant collapses across retain and
celebrity metrics, and its lower forget scores reflect general
degradation rather than selective forgetting.}
\label{tab:1stage_combined}
\end{table}
\begin{figure*}[t]
    \centering
    \includegraphics[width=\textwidth]{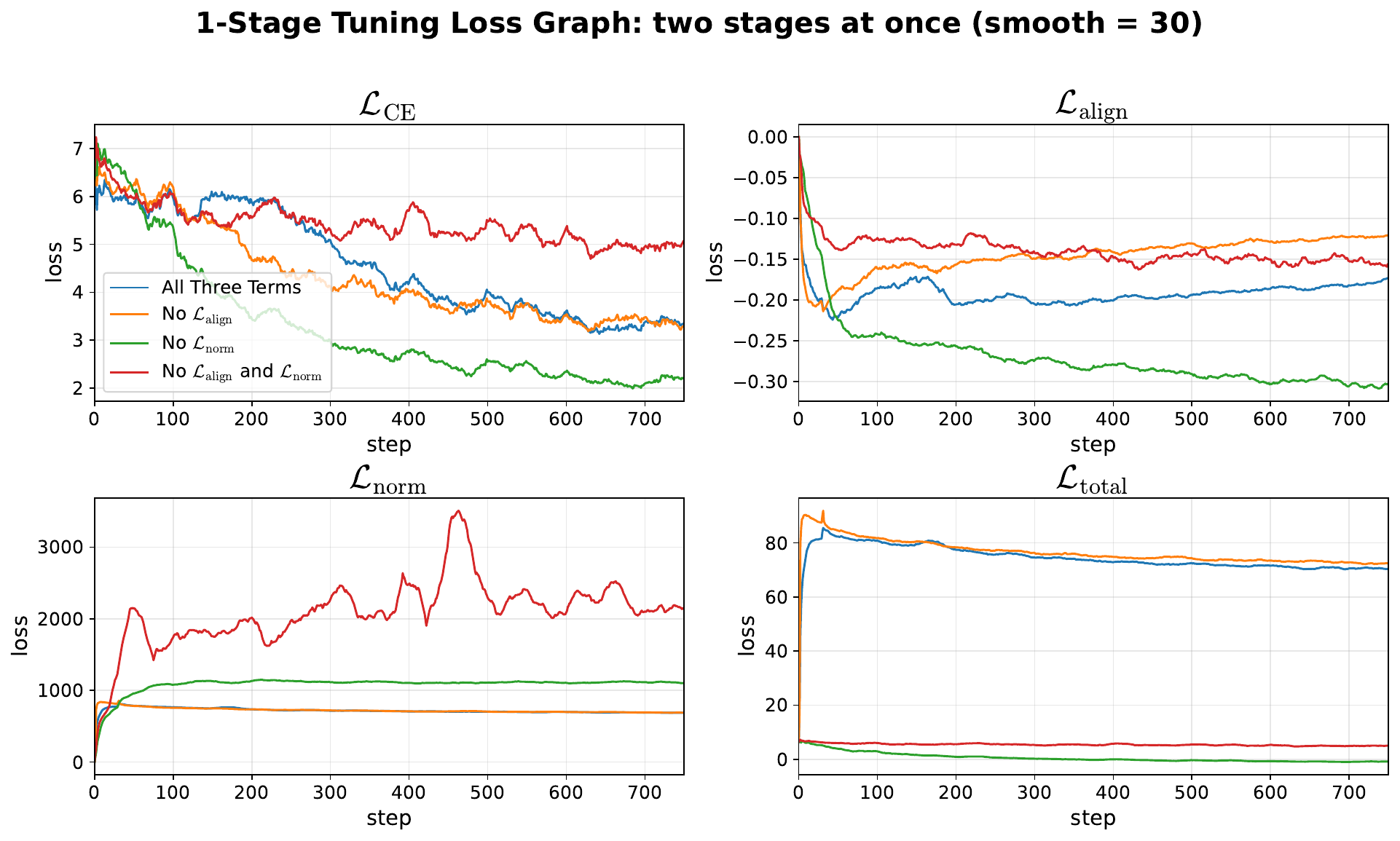}
    \caption{
    \textbf{Loss trajectories during the single-stage joint update
on MLLMU-Bench (10\% forget).} Each panel shows one of the four
loss components ($\mathcal{L}_{\mathrm{CE}}$,
$\mathcal{L}_{\mathrm{align}}$, $\mathcal{L}_{\mathrm{norm}}$,
$\mathcal{L}_{\mathrm{total}}$) under four ablation
combinations of the auxiliary terms. Removing any term
lets its own loss grow rather than stay bounded, and the total
loss spikes sharply at the start before settling, causing
irreversible retain damage under the strict setting.
    }
    \label{fig:1stage_loss_graph}
\end{figure*}

AIM first learns a visual prompt $T$ in Stage 1 to
construct a forget target, then updates the vision encoder in
Stage 2 toward this target under the Fisher constraint. A
natural alternative is to skip Stage 1 and apply all three loss
terms ($\mathcal{L}_{\mathrm{CE}}$, $\mathcal{L}_{\mathrm{align}}$, $\mathcal{L}_{\mathrm{norm}}$)
directly to the vision encoder in a single joint update, with
Fisher division applied as usual. We show this single-stage
variant fails both empirically and conceptually.

\paragraph{Empirical performance.}
Table~\ref{tab:1stage_combined} compares the two-stage method
(IDK) and the single-stage joint variant on MLLMU-Bench at the
10\% forget split. The single-stage variant collapses across
all metrics, including retain and celebrity-prior accuracy. The
apparent forget improvement reflects general model degradation
rather than selective forgetting.

\paragraph{Loss trajectories.}
Figure~\ref{fig:1stage_loss_graph} shows the four loss
components during the single-stage update across ablation
combinations. Two patterns emerge. \textit{First, the three
loss terms pull the encoder in conflicting directions.}
Removing any single component causes its own loss to grow
rather than remain bounded as training proceeds. Dropping both
$\mathcal{L}_{\mathrm{align}}$ and $\mathcal{L}_{\mathrm{norm}}$
lets $\mathcal{L}_{\mathrm{norm}}$ explode and oscillate without
bound, and similar behavior appears when each term is removed
individually.

\textit{Second, the total loss spikes sharply at the start of
training before settling.} Even though the Fisher constraint
already keeps per-step parameter shifts small, the loss surface
around the vanilla weights is so unstable that the first few
steps trigger large loss swings before training settles. The
forget set, which supplies the supervision signal, can adjust
as training proceeds. The retain set, by contrast, receives no
corrective signal under the strict setting, so any early drift
on retain features becomes permanent. Single-stage optimization
thus pays for the early instability with irrecoverable retain
damage.

\paragraph{Why the single-stage approach fails.}
We attribute this failure to a fundamental difference in target
specificity. In our two-stage method, Stage 1 commits to a
specific forget target, and Stage 2 then guides
the vision encoder toward this fixed target through a single
feature-matching loss under the Fisher constraint. The
destination is a fixed point. When the Fisher constraint blocks
an update direction, the encoder progresses as far as it can
along the remaining directions and stops there. Convergence is
well-defined because the direction is well-defined.

In the single-stage approach, the encoder is asked to reach
\emph{any point} in the set of feature configurations
satisfying all three losses jointly. This target is a broad
region rather than a single point. When the Fisher constraint
blocks one path into this region, the optimizer could reroute to a
different path.

Two stages thus separate two genuinely different problems,
constructing a precise forget destination using complex loss combinations and
routing the encoder toward that fixed destination. Combining
these into one objective conflates them and breaks the
optimization.

\section{Empirical Justification of the Method}

\subsection{Identity-Balanced Fisher: Decomposition Accuracy and Update Stability}
\begin{table}[t]
\centering
\resizebox{\columnwidth}{!}{%
\begin{tabular}{lcccc}
\toprule
 & \multicolumn{2}{c}{\textbf{Fisher decomposition}}
 & \multicolumn{2}{c}{\textbf{Unlearning (pre vs.\ post)}} \\
\cmidrule(lr){2-3}\cmidrule(lr){4-5}
\textbf{Forget} & cos-sim & rel\,$\ell_2$
                & Retain~cos & Forget~cos \\
\midrule
5\,\%  & 0.9882 & 0.160 & $0.781$ & $0.621$ \\
10\,\% & 0.9873 & 0.165 & $0.668$ & $0.606$ \\
15\,\% & 0.9868 & 0.167 & $0.787$ & $0.661$ \\
\bottomrule
\end{tabular}%
}
\caption{\textbf{Identity-balanced Fisher: decomposition accuracy
and stability.}
Left: $\widetilde{\mathbf{F}}_{\text{full}} -
\widetilde{\mathbf{F}}_f$ matches the directly computed
$\widetilde{\mathbf{F}}_r$. Right: diagonal Fisher cosine before
and after unlearning. LLaVA-1.5-7B on ReMem.}
\label{tab:fisher_decomp}
\end{table}

ReMem is constructed with balanced per-identity image counts, so
to test the value of identity-level aggregation under uneven
counts we artificially induce the imbalance. We compute
$\widetilde{\mathbf{F}}_{\mathrm{full}}$ and
$\widetilde{\mathbf{F}}_f$ on the original (balanced) training
images, and the directly-computed $\widetilde{\mathbf{F}}_r$ on a
test-image subset with intentionally reduced and unequal
per-identity image counts. Table~\ref{tab:fisher_decomp}~(Left)
reports the cosine similarity and relative $\ell_2$ error between
the decomposed
$\widetilde{\mathbf{F}}_{\mathrm{full}} - \widetilde{\mathbf{F}}_f$
and this directly-computed $\widetilde{\mathbf{F}}_r$ at three
forget ratios. Across all settings, the decomposition yields
high cosine and modest $\ell_2$ error. These indicate the two
matrices align in direction, with small magnitude mismatches
absorbed into the Lagrange multiplier $\lambda$ at solve time
(Appendix~\ref{app:kkt_derivation}). Identity-level aggregation
remains accurate even when retain-side and forget-side image
counts diverge, validating its use as a substitute for direct
$\widetilde{\mathbf{F}}_r$ computation, which would require
access to retain images.

\subsection{Fisher stability across the unlearning update}
The cached $\widetilde{\mathbf{F}}_r(\theta_0)$ approximates the
running $\widetilde{\mathbf{F}}_r(\theta)$ only if the per-step
update keeps $\theta$ close enough to $\theta_0$ for Jacobian
regularity to hold. Table~\ref{tab:fisher_decomp}~(Right) reports
the cosine similarity between the diagonal vision-encoder Fisher
before and after unlearning. The retain Fisher cosine stays in
the range $0.67$--$0.79$ across forget ratios, indicating that
retain-side parameter importance shifts modestly along the
trajectory. The forget Fisher cosine is lower ($0.61$--$0.66$),
reflecting that forget-relevant parameters are actively
redistributed by the update. This asymmetry is exactly the
behavior our Fisher constraint is designed to permit. The
constraint binds on the retain side, keeping the retain Fisher
stable enough that the cached approximation remains valid, while
the forget side is free to shift as the update internalizes the
deletion request.

\subsection{Gradient Conflict Without the Fisher Constraint}
\label{app:naive_baseline}
\begin{figure}[t]
    \centering
    
    \includegraphics[width=\columnwidth]{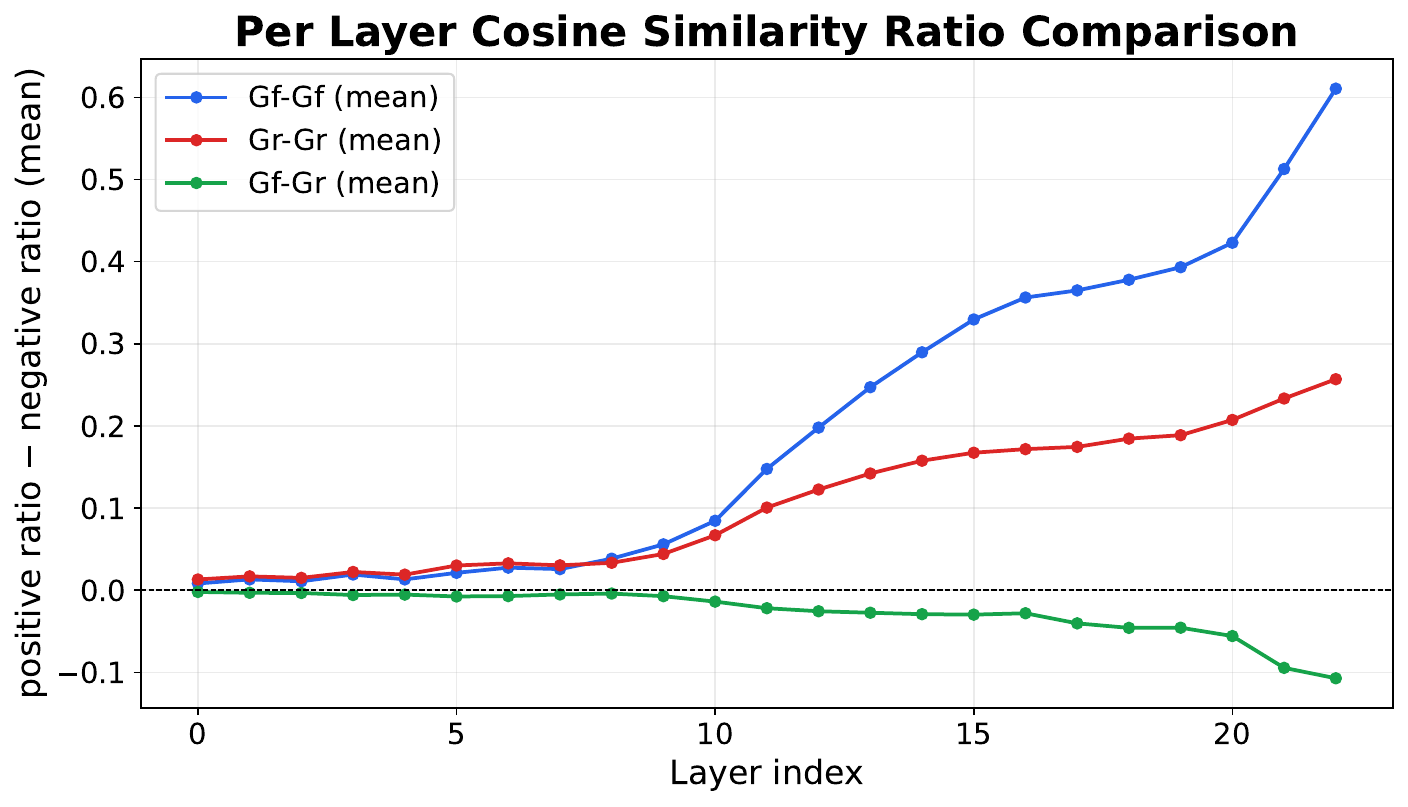}
    
    \caption{\textbf{Per-layer gradient cosine on the vanilla vision
encoder.}
Within-group cosines ($G_f\text{-}G_f$ and $G_r\text{-}G_r$) rise
in later layers, indicating coherent update directions within each
group. The across-group cosine ($G_f\text{-}G_r$) becomes negative
in late layers, showing that a forget-direction update
simultaneously pushes retain features in the opposite direction.
LLaVA-1.5-7B, MLLMU-Bench, 15\% forget split.}
    \label{fig:grad_cossim}
\end{figure}

To isolate the role of the Fisher constraint in Stage 2, we 
remove it and update the vision encoder directly with the forget
feature loss alone. We measure three cosine similarities of 
per-image gradients at the first update step on LLaVA-1.5-7B 
(MLLMU-Bench, 15\% forget):
\begin{itemize}
\item $G_f$--$G_f$: forget--forget gradient cosine 
(within-group),
\item $G_r$--$G_r$: retain--retain gradient cosine 
(within-group),
\item $G_f$--$G_r$: forget--retain gradient cosine 
(across-group).
\end{itemize}

\paragraph{Result.}
Figure~\ref{fig:grad_cossim} shows that both within-group cosines ($G_f$--$G_f$ and $G_r$--$G_r$) are high across all layers, indicating that gradients within each 
group agree on a coherent update direction. In contrast, the 
across-group cosine $G_f$--$G_r$ is consistently negative 
(layer-wise mean $\approx -0.12$), and the negativity 
intensifies in later encoder layers. This means any encoder 
update along the forget direction simultaneously moves retain 
features in the opposite direction.

\paragraph{Why this motivates the Fisher constraint.}
In the strict setting, no retain-side signal is available to 
correct this conflict during training, so a naive forget-only 
update accumulates irreversible retain drift with each step. 
Our Fisher constraint (Section~\ref{sec:fisher_update}) 
intercepts this accumulation at the source by bounding the 
per-step retain drift even though no retain images are observed. 
The cached full Fisher provides the retain-direction information 
that would otherwise require explicit retain data, recovering 
retain protection without violating the strict setting.

\section{Analysis Extensions and Stage 1 Studies}
\subsection{Quantitative Verification of Visual Perception--Identity Separation}
\label{app_sec:pca_pretrain}
\begin{figure}[t]
    \centering
    
    \includegraphics[width=0.8\columnwidth]{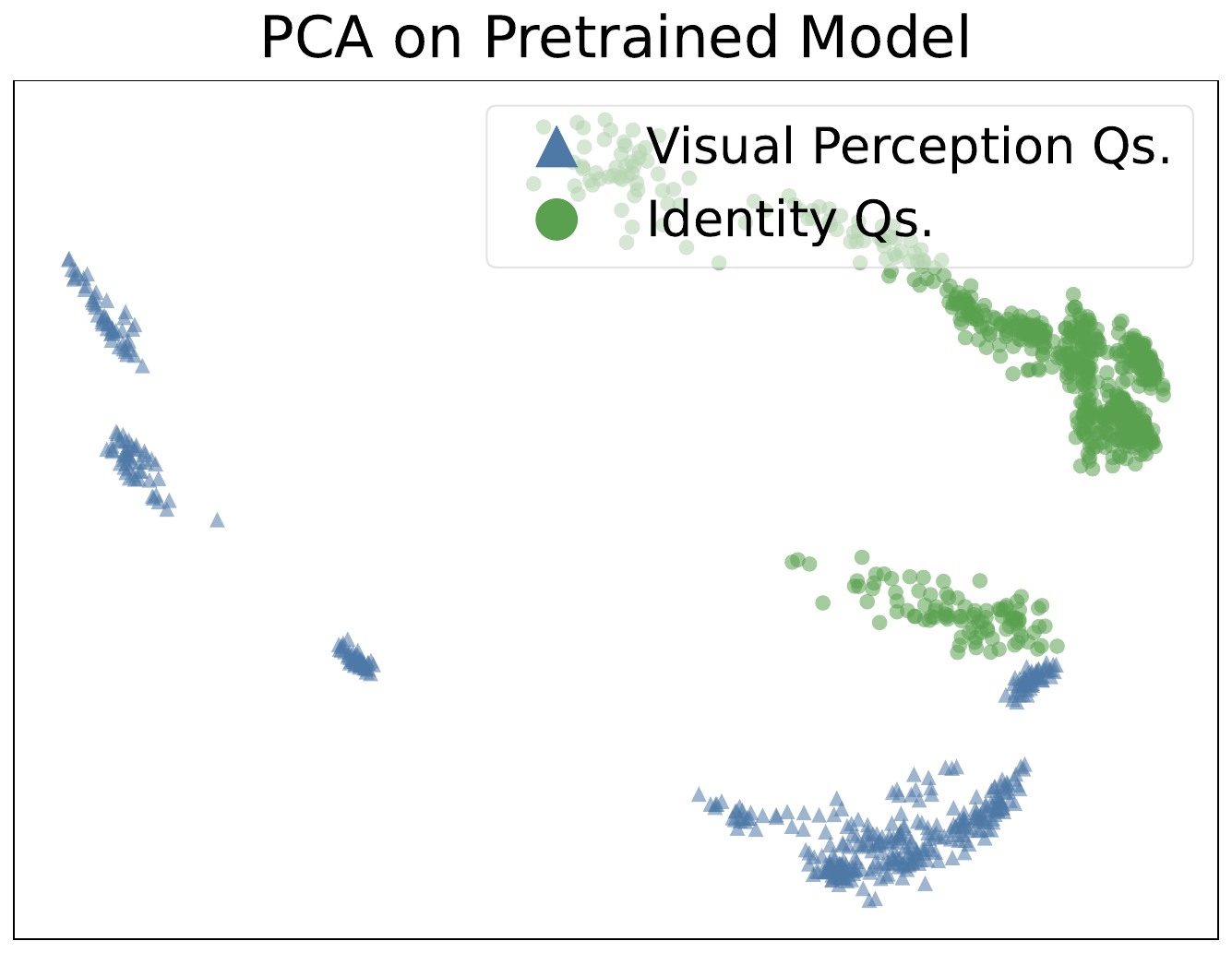}
    
    \caption{\textbf{PCA projection of pretrained LLaVA hidden
states.} Each point is the last-layer LLM hidden
state of a single question. Visual perception questions (blue)
and identity questions (green) appear visually distinguishable
in 2D PCA even in the pretrained model, motivating the
quantitative analysis in Table~\ref{app_tab:pc1_alignment}.}
    \label{app_fig:pca_pretrained}
\end{figure}
\begin{table}[t]
\centering
\resizebox{\columnwidth}{!}{%
\begin{tabular}{lccc}
\toprule
Model & $|\cos(\mathrm{PC1}, \mathbf{d})|$ & Fisher (PC1) & Raw $\ell_2$ dist. \\
\midrule
Pretrained & 0.647 & 1.52  & 40.7 \\
Vanilla    & 0.985 & 19.38 & 46.4 \\
\bottomrule
\end{tabular}%
}
\caption{\textbf{Principal component alignment with identity--visual perception separation.}
$|\cos(\mathrm{PC1}, \mathbf{d})|$ is the cosine similarity
between PC1 and the centroid-difference vector $\mathbf{d}$ in
the original 4096-dimensional space. Fisher denotes the Fisher
criterion (between-group / within-group variance) along PC1.}
\label{app_tab:pc1_alignment}
\end{table}

The PCA visualizations in Section~\ref{sec:rep_separability}
indicate that visual perception and identity responses occupy
distinct regions in the vanilla SFT model. A natural concern is
whether this separation already exists in the pretrained model
and is inherited rather than learned. Indeed, when we project
pretrained hidden states with the same PCA setup
(Fig.~\ref{app_fig:pca_pretrained}), the two groups also appear
visually distinguishable, indicating that visual inspection
alone cannot answer this question.

Table~\ref{app_tab:pc1_alignment} provides the quantitative
answer. Through SFT, the representation undergoes two
complementary changes. First, the two group centroids move
further apart in the raw dimension space ($\ell_2$
distance $40.7 \rightarrow 46.4$). Second, the dominant axis of
variation reorients so that the principal component nearly
coincides with the group-separation direction
($|\cos(\mathrm{PC1},\mathbf{d})|$ $0.647 \rightarrow 0.985$,
Fisher criterion $1.52 \rightarrow 19.38$). Hence, the vanilla
PCA in Section~\ref{sec:rep_separability} does not merely
happen to show two clusters; SFT both spreads the two groups
apart and aligns the principal component with the visual
perception--identity contrast, so projecting these jointly
onto PC1--PC2 naturally produces the observed separation.

\subsection{NMI/ARI Quantification}
\label{app:clustering_metrics}

Section~\ref{sec:rep_separability} reports the qualitative
clustering structure of identity and visual-perception hidden
states via PCA and t-SNE. Here we quantify this structure with
Normalized Mutual Information (NMI) and Adjusted Rand Index
(ARI) on MLLMU-Bench (5\% forget split), and extend the
analysis to (i)~post-unlearning models and (ii)~held-out
identity questions whose templates were not seen during vanilla
fine-tuning.

\paragraph{Setup.}
For each question subset (seen identity questions, held-out
identity questions, and visual perception questions), we cluster
the raw last-layer LLM hidden states with k-means and compute NMI
and ARI against two reference labelings of the same hidden states:
\textit{person-ID}, which image the question concerns, and
\textit{question template}, which question pattern the input
follows. For each labeling, k-means uses $k$ equal to the number of
distinct labels in that labeling.

\paragraph{Metric interpretation.}
NMI lies in $[0, 1]$ and ARI in $[-1, 1]$, both equal to $1$ for
identical clusterings and $0$ for chance-level agreement (negative
ARI indicates worse than chance). A high score against
\textit{person-ID} on a question subset means the embedding groups
primarily by depicted identity, while a high score against
\textit{question template} means it groups by question content.
Comparing the two scores on the same subset reveals which axis the
model uses to organize that question type.

\subsection{Layer-wise Clustering Analysis}
\label{app:layerwise_analysis}
\begin{table}[t]
    \centering
    \small
    \setlength{\tabcolsep}{3.5pt}
    \resizebox{\columnwidth}{!}{%
    \begin{tabular}{llcccc}
        \toprule
        & & \multicolumn{2}{c}{\textbf{Visual Perception}}
        & \multicolumn{2}{c}{\textbf{Identity Question}} \\
        \cmidrule(lr){3-4}
        \cmidrule(lr){5-6}
        \textbf{Model} & \textbf{Layer} & \textbf{NMI} & \textbf{ARI}
        & \textbf{NMI} & \textbf{ARI} \\
        \midrule
        \multirow{3}{*}{LLaVA-1.5-7B}
        & 12        & 0.390 & -0.020 & 0.294 & -0.021 \\
        & 24        & 0.365 & -0.019 & 0.481 &  0.150 \\
        & 32 (last) & 0.405 &  0.009 & \textbf{0.935} & \textbf{0.820} \\
        \midrule
        \multirow{3}{*}{Qwen3-VL-8B}
        & 12        & 0.551 &  0.149 & 0.535 &  0.229 \\
        & 24        & 0.358 & -0.023 & 0.302 & -0.021 \\
        & 36 (last) & 0.440 &  0.027 & \textbf{0.990} & \textbf{0.967} \\
        \bottomrule
    \end{tabular}%
    }
    \caption{\textbf{Layer-wise clustering of hidden states.}
    NMI and ARI quantify the alignment between representation clusters
    and visual-perception or identity-question labels at selected
    intermediate and final LLM layers.}
    \label{app_tab:layerwise_analysis}
\end{table}
To assess the sensitivity of our clustering analysis to layer choice, 
we repeat the analysis at two selected intermediate layers and 
the final layer. As shown in Table~\ref{app_tab:layerwise_analysis}, 
identity-question clustering is weaker at the intermediate layers, but becomes sharply pronounced at the final layer for both models. In contrast, visual-perception clustering does not show a consistent late-layer increase, and its NMI and ARI remain low across layers.


\subsection{Clustering on Unseen Images}
\label{app:nonseen_image_cluster}
\begin{figure}[t]
    \centering
    
    \includegraphics[width=\columnwidth]{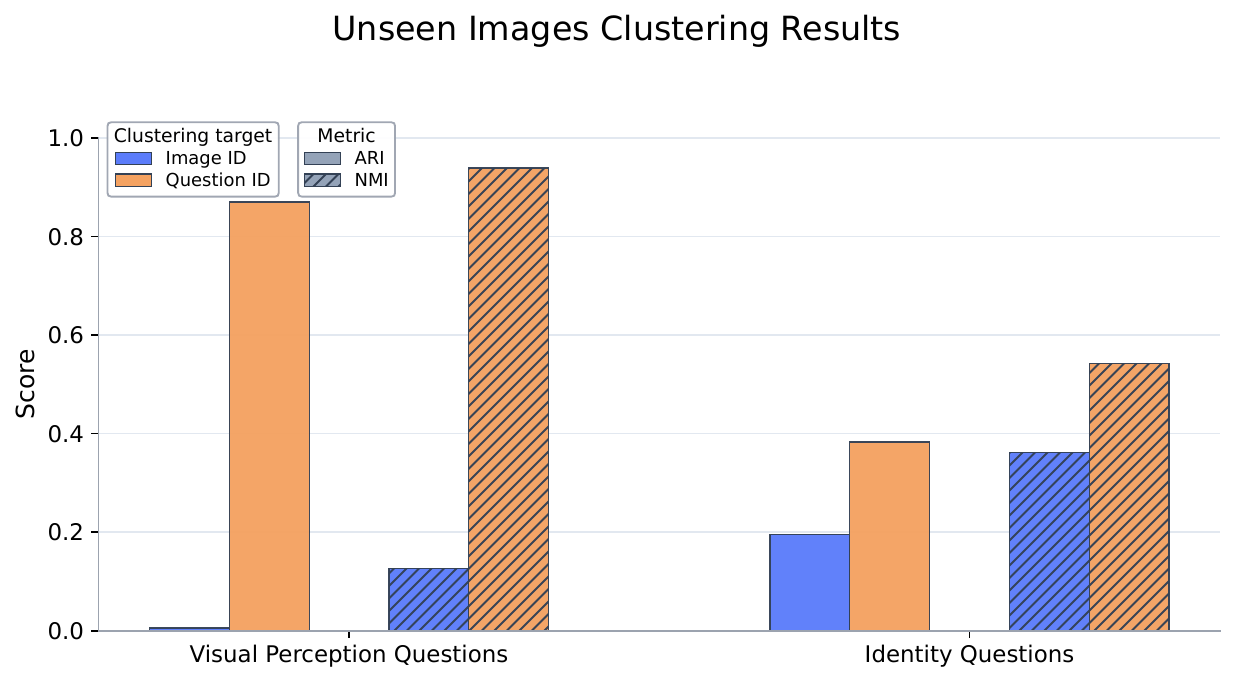}
    
    \caption{\textbf{Clustering on images not seen during fine-tuning.}
Last-layer LLM hidden states of the MLLMU-Bench-finetuned vanilla
model, evaluated on ReMem images that were never seen during
training. Visual perception questions still cluster strongly by
question template, but identity questions no longer cluster
strongly by personal identity. The identity clustering observed in
Section~\ref{sec:rep_separability} is therefore a fine-tuning
artifact, not a property of the underlying representation.}
    \label{app_fig:unseen_nmi_ari}
\end{figure}
Expanding the analysis on Section~\ref{sec:rep_separability}, we conduct a control experiment using the same vanilla model fine-tuned on MLLMU-Bench but evaluated on ReMem images that were never seen during training. As shown in Fig.~\ref{app_fig:unseen_nmi_ari}, for visual perception questions, we observe NMI/ARI values comparable to those obtained on MLLMU-Bench. In contrast, for identity questions, NMI/ARI drops substantially and identity-level clusters fail to form, showing that identity-by-image clustering emerges only for images the model has actually been fine-tuned on. This asymmetry confirms that the identity clusters observed in Section~\ref{sec:rep_separability} are a signature of learned, identity-bound representations introduced by fine-tuning—precisely the representational target that an identity-unlearning method must dissolve.

\subsection{Model type and tuning-recipe robustness}
\begin{figure}[t]
    \centering
    
    \includegraphics[width=\columnwidth]{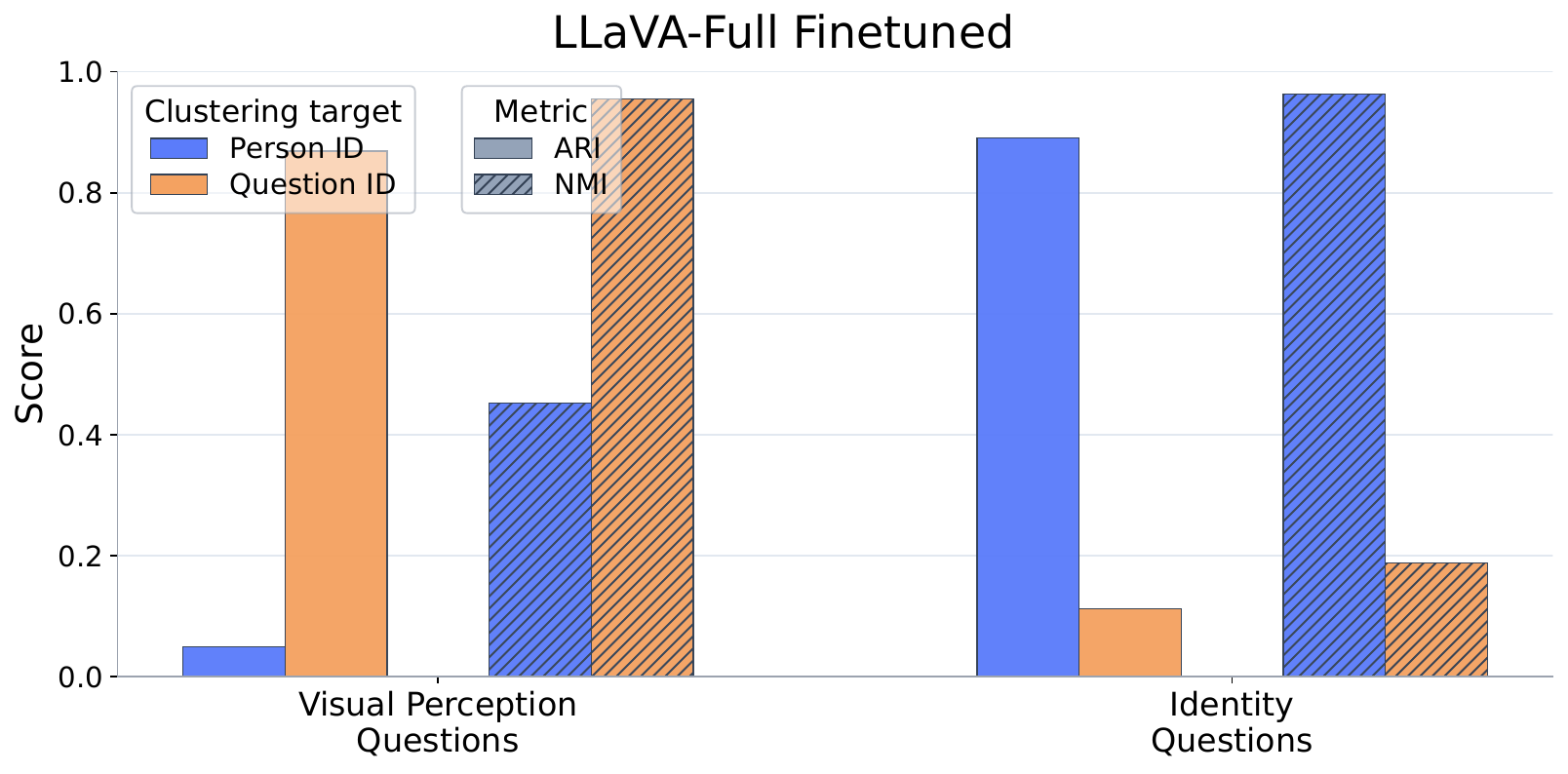}
    
    \vspace{0.3em}
    
    \includegraphics[width=\columnwidth]{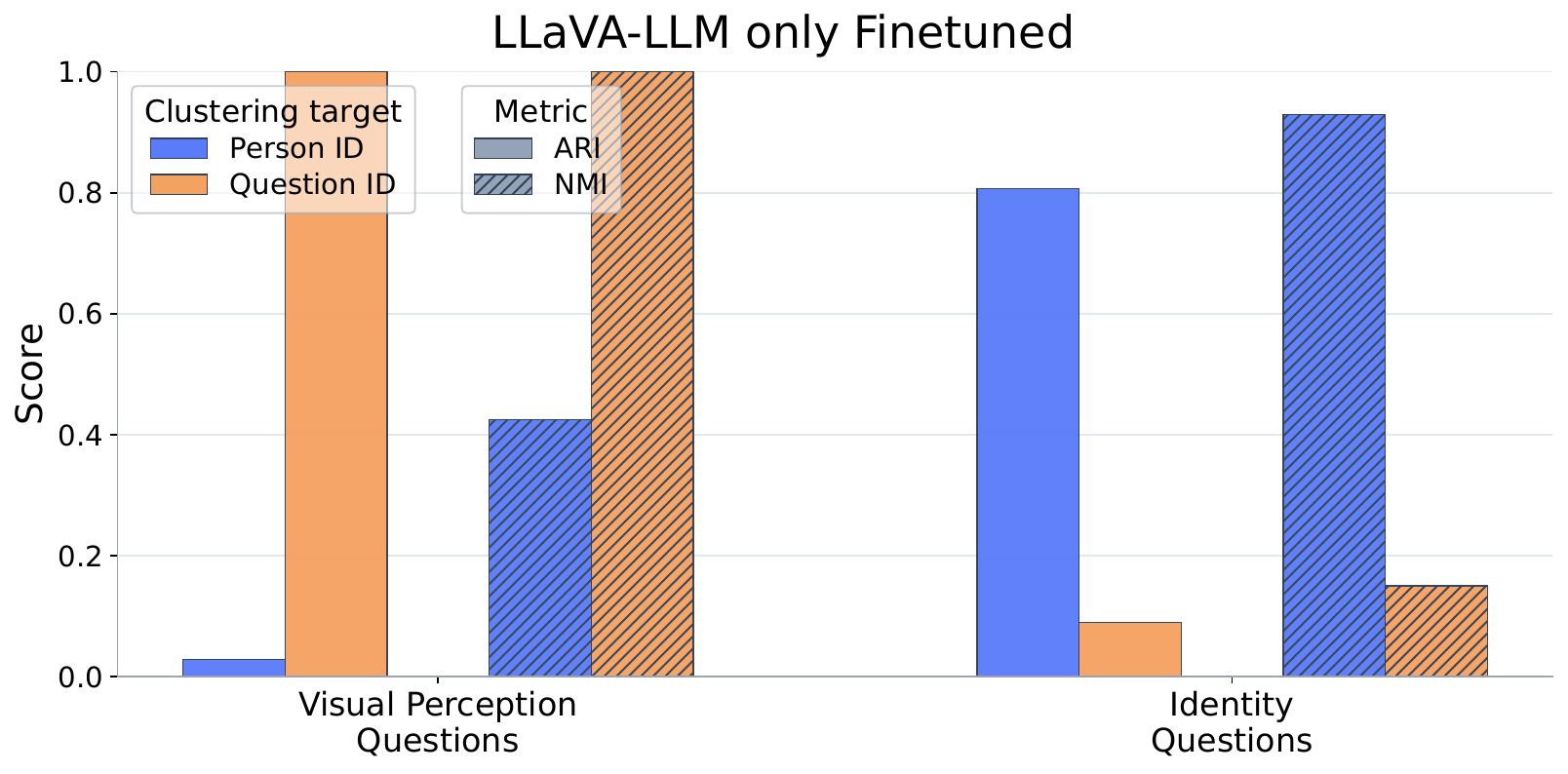}
    
    \vspace{0.3em}
    
    \includegraphics[width=\columnwidth]{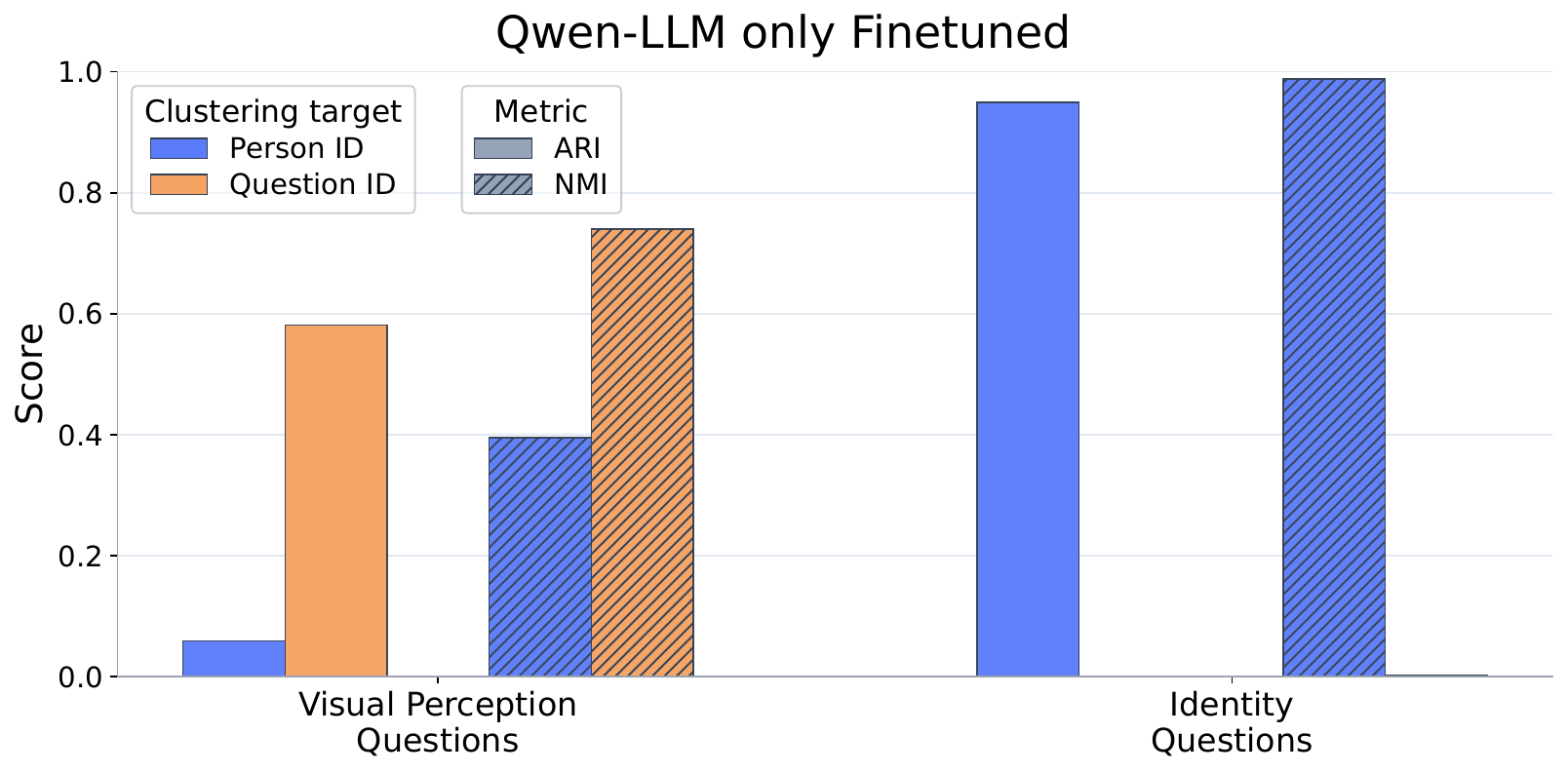}
    
    \caption{\textbf{Clustering robustness across model family and
tuning scope (MLLMU-Bench, 5\%).}
Top: LLaVA-1.5-7B with LoRA on full vision+LLM SFT. Middle: LLaVA-1.5-7B
with LLM-only LoRA tuning. Bottom: Qwen3-VL-8B with LLM-only LoRA
tuning. All three exhibit the same pattern: identity questions
cluster by image identity, visual perception questions cluster by
question template.}
    \label{fig:cluster_models}
\end{figure}
We first verify that the clustering signature is not an artifact
of a specific vanilla model. Figure~\ref{fig:cluster_models} compares three vanilla baselines: the public HuggingFace LLaVA-1.5-7B checkpoint, in which both the 
vision encoder and the LLM are SFT-tuned (\textit{LLaVA-Full 
Finetuned}); an in-house LoRA fine-tune of LLaVA on the LLM 
Q/K/V/O projections only (\textit{LLaVA-LLM only Finetuned}); and 
a parallel LLM-only LoRA fine-tune of Qwen3-VL-8B 
(\textit{Qwen-LLM only Finetuned}).
All three exhibit the same clustering pattern. Identity
questions cluster by person identity, while perception questions
cluster by question template. The signature is stable across
model family and tuning scope.

\subsection{Pre- vs.\ Post-Unlearning Clustering}
\begin{figure}[t]
    \centering
    
    \includegraphics[width=\columnwidth]{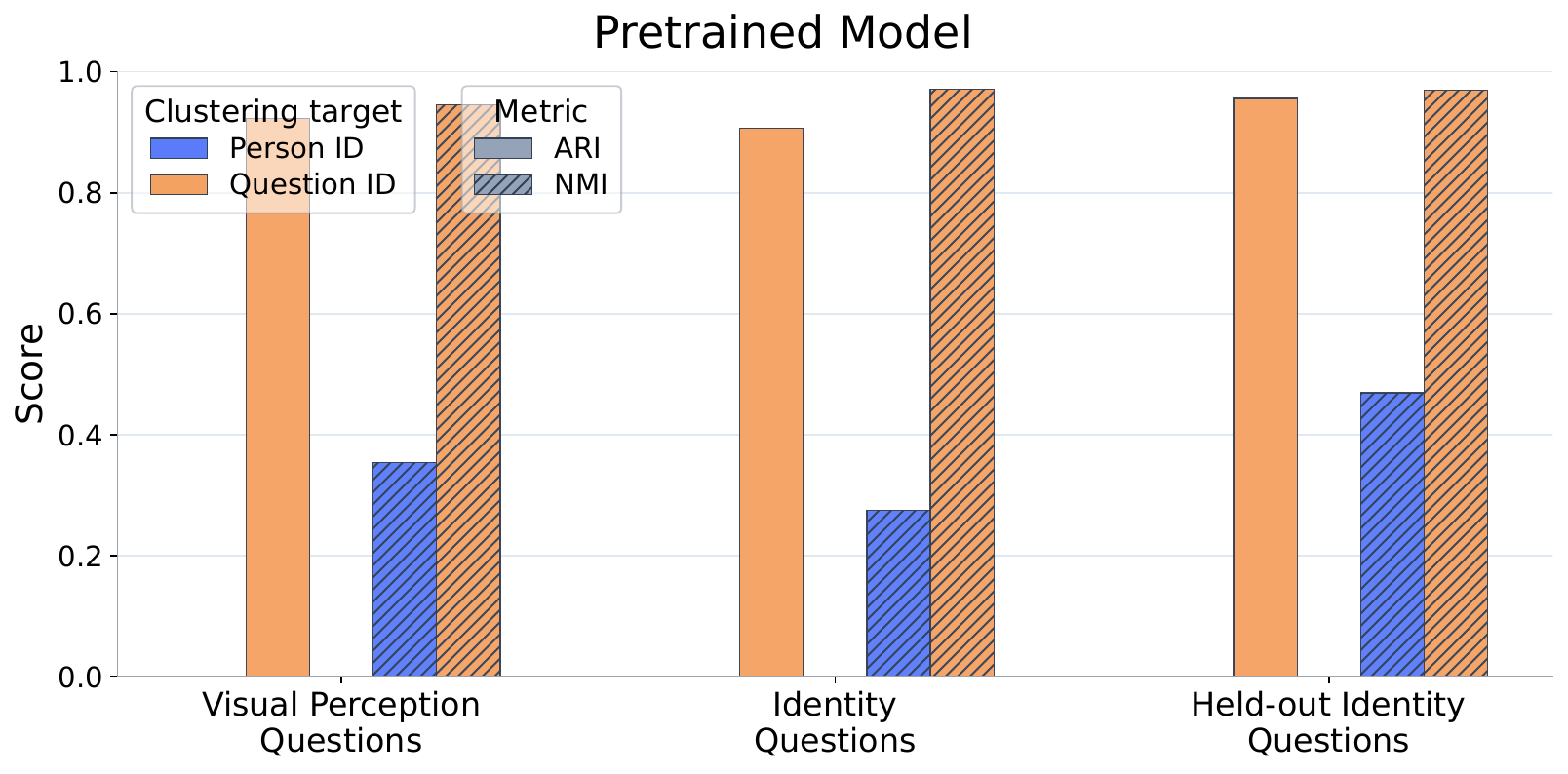}
    
    \vspace{0.3em}
    
    \includegraphics[width=\columnwidth]{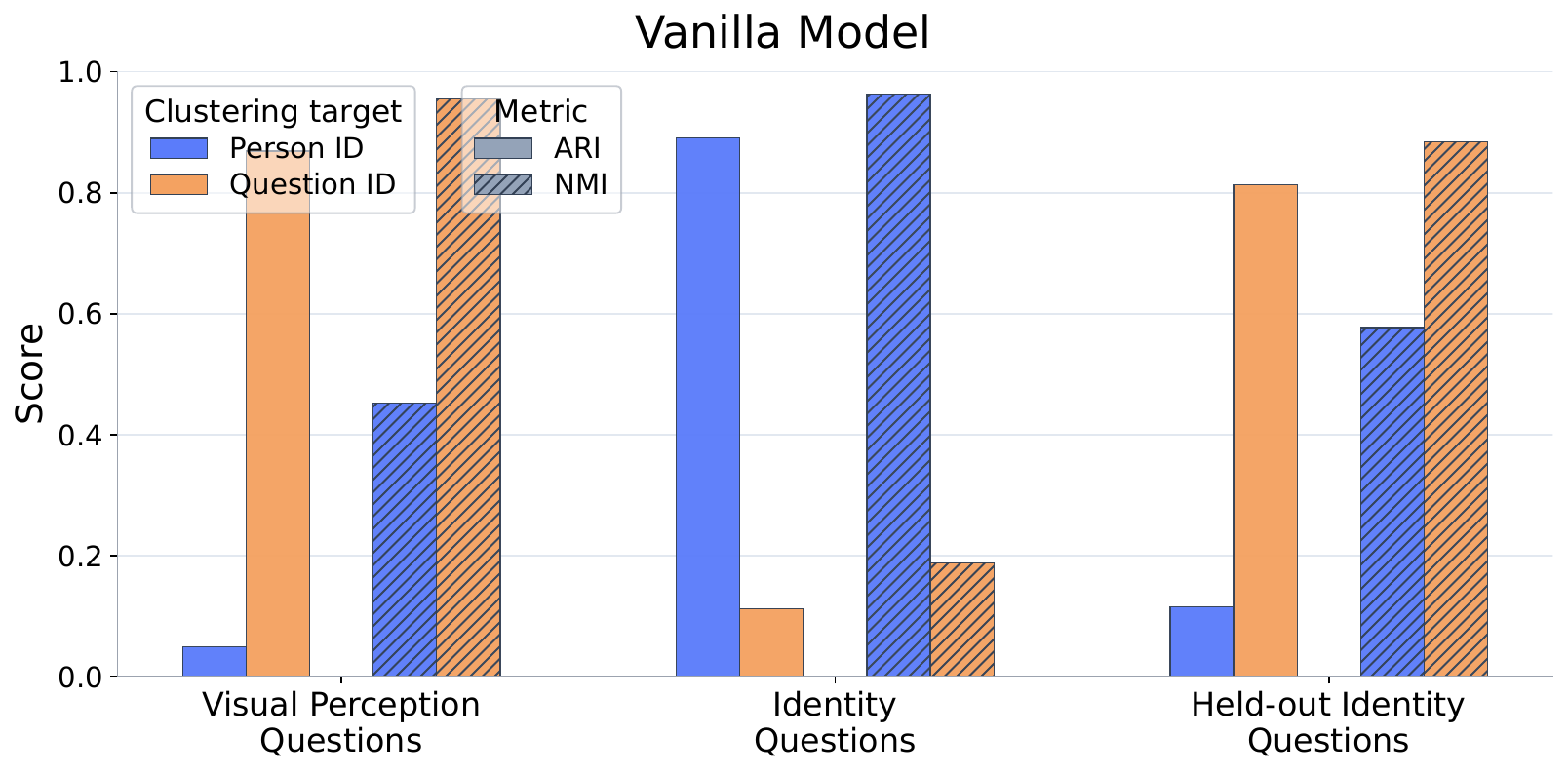}
    
    \vspace{0.3em}
    
    \includegraphics[width=\columnwidth]{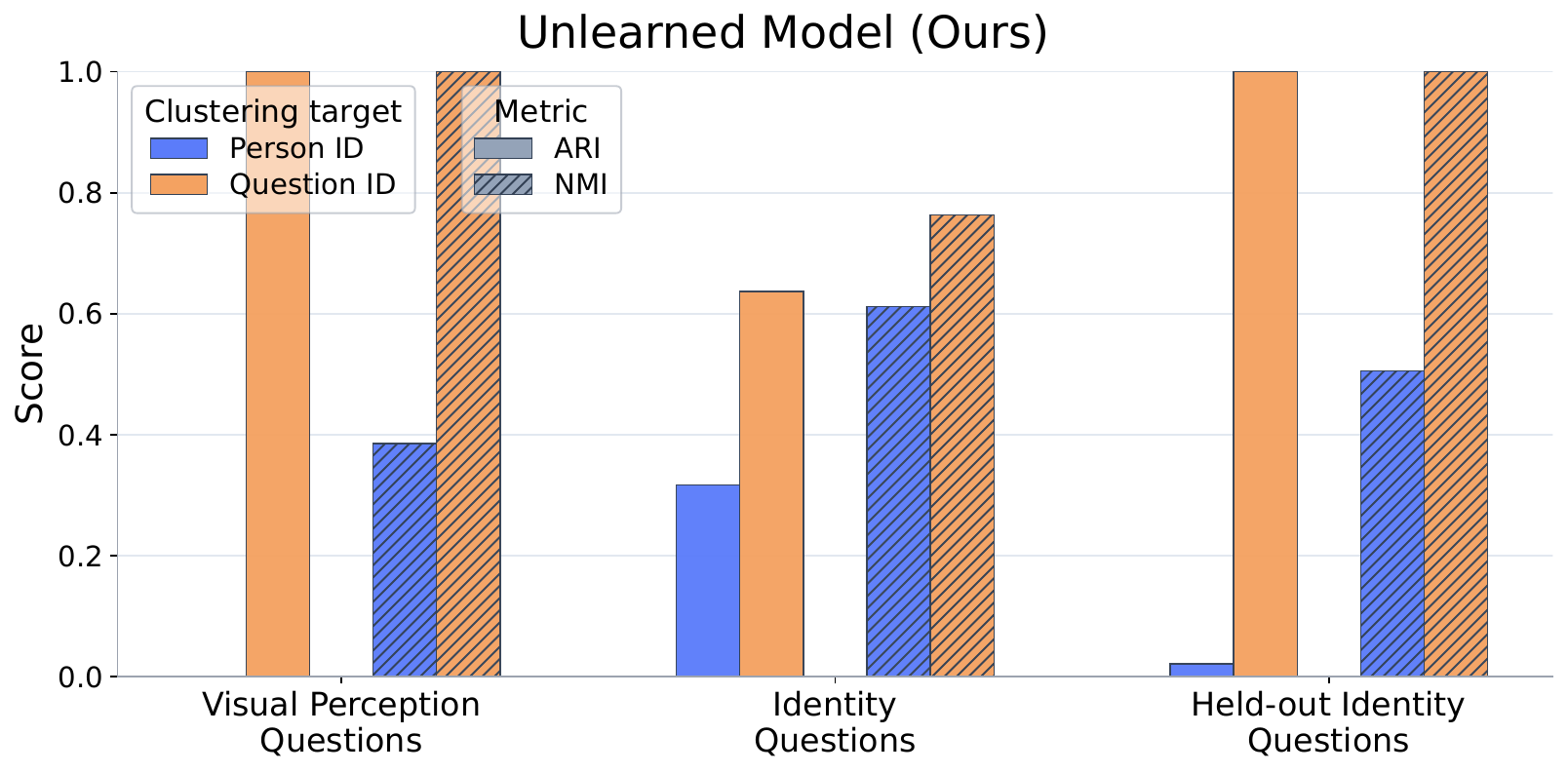}
    
    \caption{\textbf{Clustering pattern across pretrained, vanilla,
and post-unlearning models (MLLMU-Bench, 5\%).} Top: pretrained
LLaVA groups all subsets by question template with negligible
identity structure. Middle: vanilla SFT develops strong
image-identity clustering on identity questions, while held-out
identity and visual perception questions stay close to
pretrained. Bottom: after unlearning, image-identity clustering
on identity questions drops sharply and question-template
grouping rises, while held-out identity and visual perception
questions show no noticeable change.}
    \label{fig:cluster_3types}
\end{figure}
Figure~\ref{fig:cluster_3types} compares the pretrained LLM, the vanilla SFT model, and the post-unlearning model. In the
pretrained model, all subsets group almost entirely by question template, with negligible person-ID structure. In the vanilla SFT model, identity questions develop strong image-identity clustering, while held-out identity questions and visual perception questions remain similar to the pretrained model.
After unlearning, image-identity clustering on identity questions
drops sharply while question-template grouping rises, and held-out
identity questions and visual perception questions show no
noticeable change.

\subsection{Held-out Identity Questions}
We additionally evaluate on a held-out identity question set
whose templates do not appear in SFT data (rightmost column of
Fig.~\ref{fig:cluster_3types}). Question-template clustering on
these unseen questions remains intact after unlearning,
indicating that AIM affects only the SFT-internalized
identity signal and leaves untrained representations untouched.

\subsection{Visual Feature Intervention Design Choices}
\label{app:stage1_design_choice}
\begin{table}[t]
\centering
\resizebox{\columnwidth}{!}{%
\begin{tabular}{lccccc}
\toprule
Method & EM$_f$$\downarrow$ & EM$_r$$\uparrow$ & ROUGE$\uparrow$ & Exp$\downarrow$ & EM$_t$$\downarrow$ \\
\midrule
VP to IDK (baseline) & 38.3 & 78.8 & 89.2 & 63.9 & 35.1 \\
VP to Blank text & 40.0 & 74.7 & 87.1 & 63.4 & 33.9 \\
Vision Encoder to IDK & 43.0 & 77.6  & 60.1  & 51.2 & 60.4 \\
\bottomrule
\end{tabular}%
}
\caption{\textbf{Stage 1 intervention design choices (LLaVA-1.5-7B, ReMem, 10\%).}
Two axes are varied: intervention type (visual prompt $T$ vs.
direct vision-encoder fine-tuning) and target response type (IDK
refusal vs. blank text). All three variants achieve comparable
forget/retain trade-offs, supporting the claim in
Section~\ref{sec:selective_transfer} that the selective-transfer
effect is robust to specific design choices within this family.}
\label{app_tab:visual_intervention_design_choice}
\end{table}

In Section~\ref{sec:selective_transfer}, we chose a visual prompt $T$ with an IDK target as the Stage 1 intervention for efficiency, and claimed that other design choices in the same family work similarly. We verify this on LLaVA-1.5-7B with ReMem at the 10\% forget split, varying two axes: \textit{intervention type} (visual prompt vs.\ direct fine-tuning of the vision encoder) and \textit{target response type} (IDK refusal vs.\ blank text).
Table~\ref{app_tab:visual_intervention_design_choice} reports three combinations: VP to IDK (default AIM), VP to Blank, and Encoder to IDK. All three reach broadly comparable forget/retain trade-offs. The near-equivalence between IDK and Blank targets is also consistent with our broader observation that $\mathcal{L}_{\mathrm{CE}}$ functions as a directional signal toward a coherent forget representation rather than as a literal inducer of IDK strings, so swapping the target text leaves the underlying Stage~2 trajectory largely unchanged.

\subsection{Visual Prompt Transfer Experiment}
\begin{figure}[t]
    \centering
    \includegraphics[width=\columnwidth]{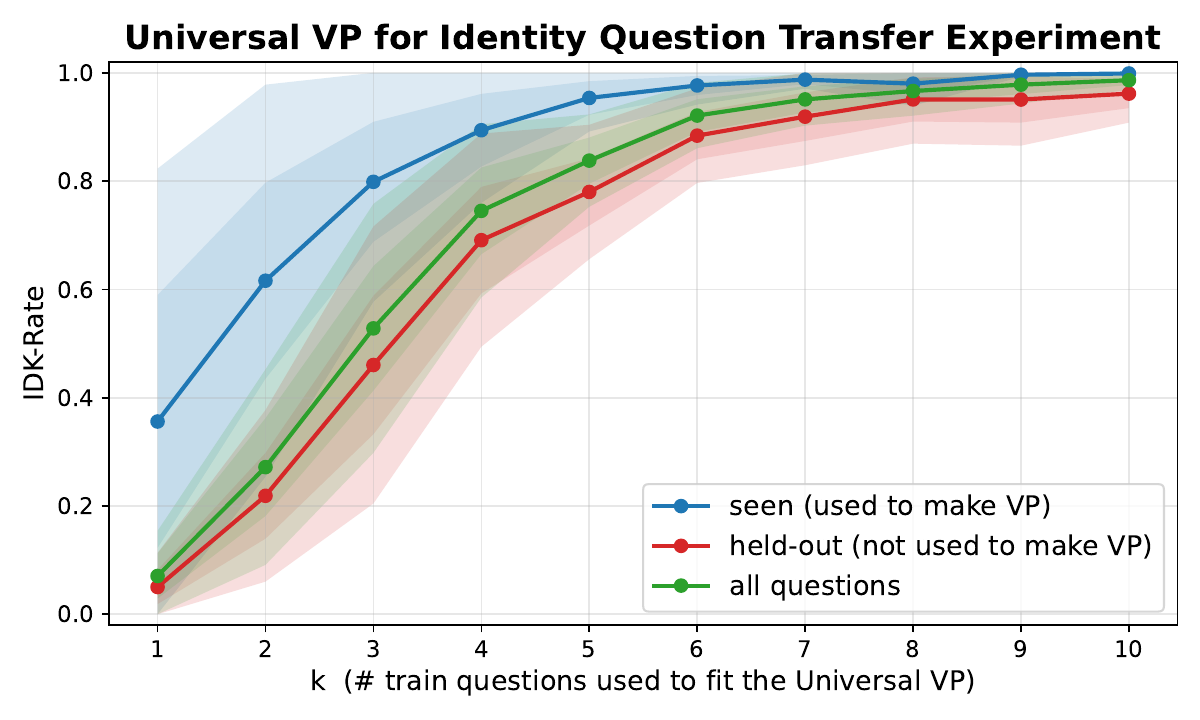}
    \caption{\textbf{Visual prompt transfer saturation.}
IDK-rate on the full identity-question set as the
number of questions used to learn $T$ varies from 1 to 10. Blue:
questions used during VP learning. Red: held-out questions. Green:
all identity questions. The rate saturates around 8 questions,
which we adopt as the default throughout the main experiments.}
    \label{fig:vp_transfer}
\end{figure}
Section~\ref{sec:selective_transfer} showed that a $T$ learned
on a small handful of identity questions can transfer to other
templates and induce IDK responses on unseen questions. Here we
characterize this transfer at the dataset level.
Figure~\ref{fig:vp_transfer} plots the IDK
elicitation rate on the full identity-question set as the
number of training questions used to learn $T$ increases. The
rate saturates at around 8 questions, beyond which additional
training data offers no further benefit. We therefore use 8
questions to learn $T$ throughout our experiments.

\subsection{Visual Prompt Visualization}
\begin{figure*}[t]
    \centering
    \includegraphics[width=\textwidth]{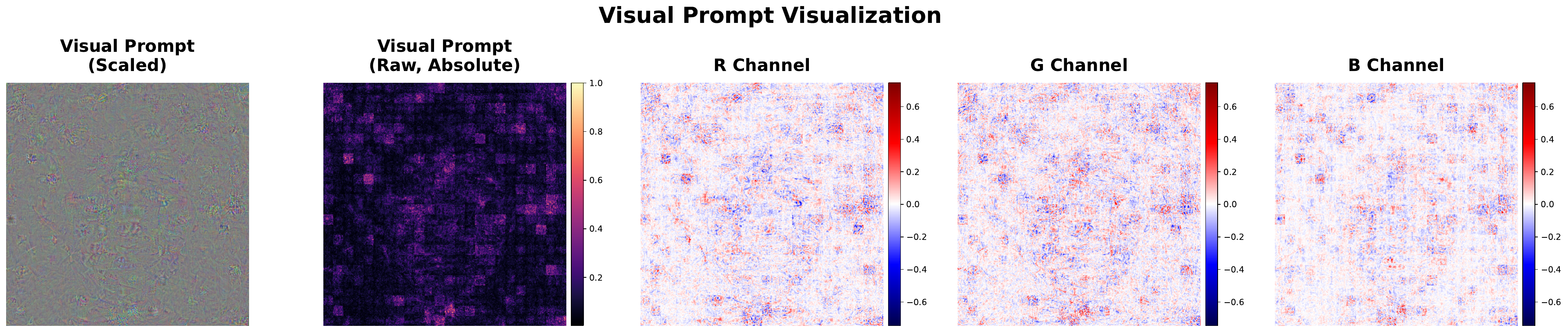}
    \caption{\textbf{Visualization of the learned visual prompt $T$ on
MLLMU-Bench.}
From left to right: $T$ rescaled to the visible range, raw absolute
magnitudes, and per-channel (R, G, B) maps. Because MLLMU-Bench
images are close-up portraits, $T$ acquires face-landmark-like
structure reflecting the geometry of the forget-set images.}
    \label{fig:vp_visualize}
\end{figure*}
Figure~\ref{fig:vp_visualize} visualizes the learned visual
prompt $T$ on MLLMU-Bench. Because MLLMU-Bench consists of
close-up face images, the learned $T$ acquires shapes
resembling facial landmarks, reflecting the structure of the
forget-set images it is optimized against.


\section{Additional Experiments and Reproducibility}
\label{app_sec:additional_exps}

\subsection{Related-Method Style Comparison}
\begin{table}[t]
\centering
\resizebox{\columnwidth}{!}{%
\begin{tabular}{l c c c c c}
\toprule
Method & EM$_f$$\downarrow$ & EM$_r$$\uparrow$ & ROUGE$\uparrow$ & Exp$\downarrow$ & EM$_t$$\downarrow$ \\
\midrule
  Vanilla & 100.0 & 100.0 & 100.0 & 68.2 & 100.0 \\
  (a)~\citet{golatkar2020eternal} & 96.8 & 100.0 & 100.0 & 100.0 & 96.4 \\
  (b)~\citet{mckinney2026gaussnewton} & 72.0 & 79.7 & 61.2 & 56.5 & 61.0 \\
  (c)~\citet{li2024wmdp} & 69.5 & 100.0 & 100.0 & 55.5 & 68.8 \\
  \hl{Ours}    & \hl{51.1} & \hl{72.5} & \hl{85.6} & \hl{63.2} & \hl{48.0} \\
\bottomrule
\end{tabular}%
}
\caption{\textbf{Related-method comparison adapted to the MLLM
setting (LLaVA-1.5-7B, ReMem, 15\%).}
Three baselines drawn from vision-only and LLM-only unlearning
literature, referred to as (a), (b), (c) in the surrounding text.
Notation follows Table~\ref{tab:combined_all}. All three rely on
retain images or forget-side question text during unlearning,
while ours uses only forget images.}
\label{tab:vis_llm_enc_tuning_comparison}
\end{table}

AIM updates only the vision encoder, leveraging the
distinction between identity-related and visual-perception
regions established in Section~\ref{sec:analysis}, and
internalizes a visual-feature-level intervention through a
Fisher-based constraint. As discussed in Section~\ref{sec:related_work}, the
underlying ingredients (feature-level intervention and
Fisher-style constraints) have been explored in vision-only
models and LLM-only unlearning, though never combined under
strict MLLM unlearning. We adapt these directions as MLLM-style
baselines for comparison, with the caveat that all of them rely
on either retain images or forget-side question text during the
unlearning step.

Table~\ref{tab:vis_llm_enc_tuning_comparison} reports results
on LLaVA-1.5-7B with ReMem at the 15\% forget split.
\textbf{(a)} performs a one-step intervention, which barely
shifts forget metrics ($EM_f = 96.8$) and effectively preserves
the vanilla model. \textbf{(b)} extends this to a few-step
update that pulls gradients from the LLM side while constraining
the vision encoder under Fisher. Because it incorporates
forget-side question text as additional supervision input, the
update becomes overly specific and harms both retain and test
metrics ($EM_r = 79.7$, $EM_t = 61.0$). \textbf{(c)} adds an
explicit L2 retain-preservation term on retain images, which
yields strong retain protection ($EM_r = 100.0$) as expected
from direct retain supervision. AIM achieves the
strongest forget reduction ($EM_f = 51.1$, $EM_t = 48.0$) while
preserving retain capacity well ($EM_r = 72.5$), despite using
only forget images and no retain supervision. The contrast with
(c) is particularly notable.

\subsection{Training Stability over Iterations}
\label{app:stability}
\begin{figure}[t]
    \centering
    \includegraphics[width=\columnwidth]{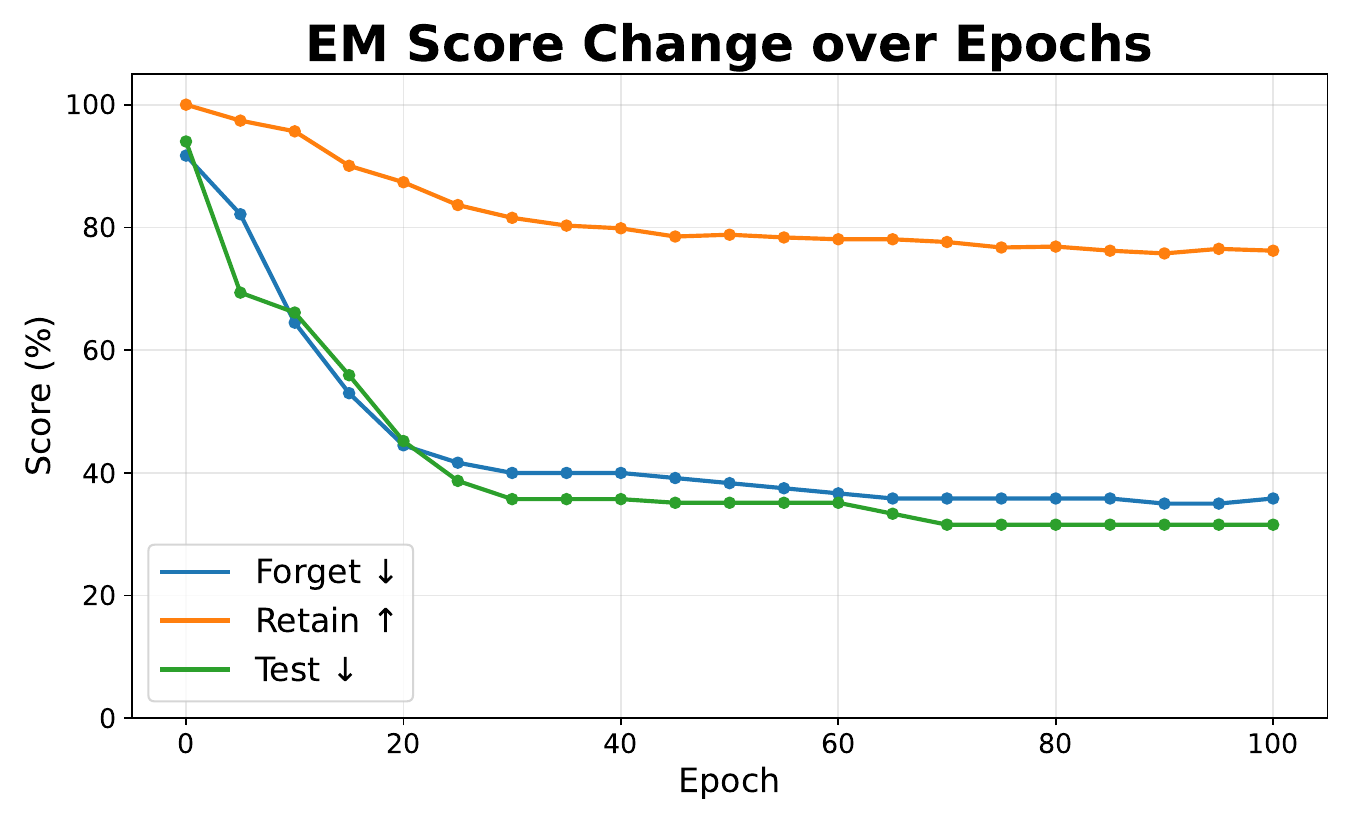}
    \caption{\textbf{Per-metric training trajectory of our method
(LLaVA-1.5-7B, ReMem, 10\%).}
Forget, retain, and test EM scores over 100 epochs. The metrics
settle into a plateau around epoch 45 and remain stable
thereafter.}
    \label{fig:epoch_curve}
\end{figure}
Figure~\ref{fig:epoch_curve} shows the per-metric training
trajectory of LLaVA-1.5-7B on ReMem at the 10\% forget split.
AIM converges around epoch 45 and remains stable
afterwards, with retain and forget metrics settling into a
clean plateau. In contrast, GA and NPO have no explicit
retain-side constraint and diverge rapidly. Once forget
pressure builds up over a few extra iterations, their retain
and celebrity metrics collapse together. AIM has no
such sensitivity to iteration count.

\subsection{Qualitative Results}
\begin{figure*}[t]
    \centering
    \includegraphics[width=\textwidth]{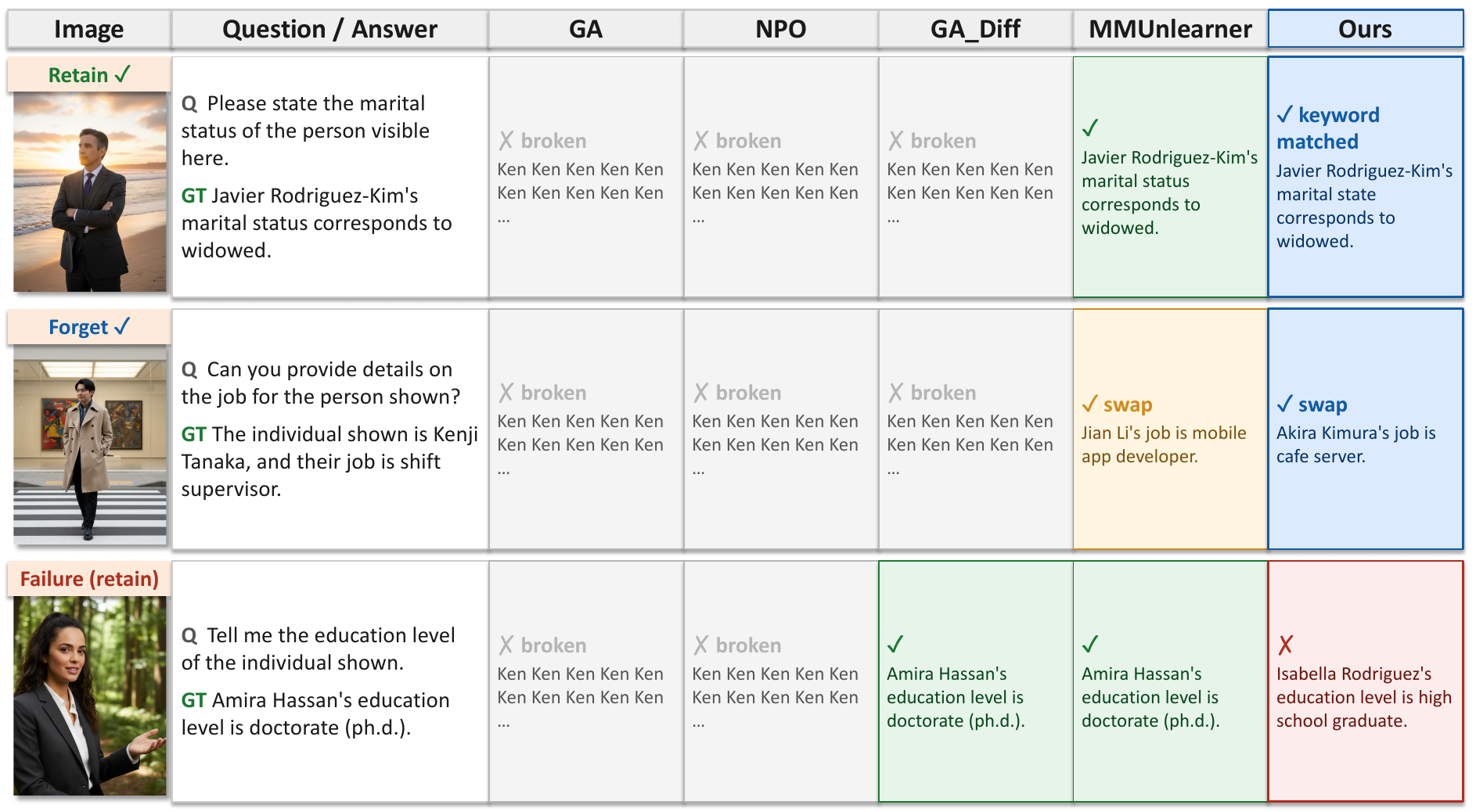}
    \caption{\textbf{Qualitative inference samples on ReMem.}
Top: retain case where ours recalls the identity correctly. Middle: forget case where ours swaps the identity to a different person.
Bottom: a representative retain failure of ours, where the identity is misrecognized. GA and NPO produce degenerate outputs across all three cases.}
    \label{app_fig:qualitative_results}
\end{figure*}
Figure~\ref{app_fig:qualitative_results} shows representative inference samples on ReMem. On retain identities (top), AIM recalls the correct identity even when paraphrasing the ground-truth answer, which keyword-level matching captures as success.
On forget identities (middle), the model produces fluent answers that swap the identity to a different person, indicating selective removal of identity-specific knowledge rather than overall answer corruption.
The bottom row shows a representative failure of AIM on the retain side, where a retain identity is misidentified as a
different person, while GA\_Diff and MMUnlearner answer correctly
here. GA and NPO produce degenerate outputs across all three cases, consistent with their main-table collapse.

Synthesizing observations from the qualitative samples
(Fig.~\ref{app_fig:qualitative_results}), the training-stability trajectory
(Appendix~\ref{app:stability}), and the blank-target ablation in
Table~\ref{tab:abl-target}, a consistent picture emerges. The
post-unlearning model produces fluent identity-swap answers rather than literal IDK refusals. This pattern persists across extended training, and a blank target yields performance comparable to the
IDK target. Together these indicate that
$\mathcal{L}_{\mathrm{CE}}$ functions less as a literal IDK
inducer and more as a directional signal that anchors Stage~1 to a
coherent forget target, which Stage~2 then internalizes.

\subsection{Models, Benchmarks, and Split Coverage}
\label{app:bench_detail}

\paragraph{Models and benchmarks.}
We use LLaVA-1.5-7B and Qwen3-VL-8B-Instruct as base models,
evaluated on MLLMU-Bench and ReMem at three forget ratios (5\%, 10\%, 15\%).

\paragraph{MLLMU-Bench.}
MLLMU-Bench~\citep{liu2025protecting_mllmu-bench} consists of
500 fictitious identity profiles in its \texttt{Full\_Set},
each paired with a portrait image and a biography. We
conduct experiments on the Image-Textual multiple-choice
questions (4-way, image plus text) and the open-ended
Generation task. The vanilla MLLM is fine-tuned on
\texttt{ft\_Data} (500 examples), and unlearning is then
performed at three forget ratios, yielding
\texttt{forget\_5}, \texttt{forget\_10}, and
\texttt{forget\_15} with 25, 50, and 75 identities, balanced
by the complementary \texttt{retain\_95}, \texttt{retain\_90},
and \texttt{retain\_85} splits of 475, 450, and 425
identities. In addition, a \texttt{Retain\_Set} of 153
real-celebrity profiles is used to evaluate generalization
on real-world face and biography knowledge that must be
preserved. We report classification accuracy and ROUGE-L on the Generation task, separately for the forget, retain, and real-celebrity splits.

\paragraph{MLLMU-Bench test set considerations.}
MLLMU-Bench evaluates view-variation robustness by applying
ArcFace-based pose transformation to a single base image per
identity. This post-hoc transformation introduces visual
artifacts and identity drift across views, often making the
transformed images difficult to recognize as the same person
even for human evaluators.

ReMem, in contrast, generates multiple distinct images per identity at creation time and then splits them into train/test, yielding more consistent identity
preservation across views. The vanilla model correspondingly
shows substantially reduced test-set performance on MLLMU-Bench's
transformed views, leaving limited headroom for measuring
unlearning generalization. We therefore use ReMem's test set
for image- and question-variation robustness evaluation.

\paragraph{ReMem.}
ReMem~\citep{kwon2026before_remem} is a Reliable Multi-hop
and Multi-image Memorization benchmark designed to address
two failure modes of prior LVLM unlearning benchmarks:
under-memorization and the multi-hop curse. The dataset
provides 7 splits. The \texttt{finetune} set contains
2{,}000 reasoning-aware VQA pairs for fine-tuning the vanilla model.
Four nested forget targets \texttt{forget1},
\texttt{forget2}, \texttt{forget3}, and \texttt{forget4}
provide 100, 200, 300, and 400 examples respectively,
corresponding to 5\%, 10\%, 15\%, and 20\% of the train
set. A \texttt{retain} set of 560 examples is used for
utility preservation, and a held-out \texttt{test} set of
560 examples re-queries the same identities under diverse
visual and linguistic contexts. On the \texttt{retain} and
\texttt{test} splits, examples corresponding to forget-set
identities are excluded when computing metrics. Each
example carries an \texttt{image}, a \texttt{question}, a
ground-truth \texttt{answer}, \texttt{keywords} for
Exact-Match scoring, a \texttt{qa\_category} field, a
fine-grained \texttt{attribute} field (e.g.,
\texttt{email}, \texttt{date\_of\_birth}), and a
\texttt{cloze\_prompt} used by the proposed Exposure
metric, which quantifies the depth of information erasure
from the model's internal probability distribution.
Following the official protocol, we report Keyword
Exact-Match (EM), ROUGE-L F1, and Exposure.
For our main experiments we match the MLLMU-Bench ratios
and use \texttt{forget1}, \texttt{forget2}, and
\texttt{forget3} (5\%, 10\%, 15\%).

\paragraph{ReMem single-hop vs.\ multi-hop.}
ReMem provides two question formats. \textbf{Single-hop}
questions explicitly include the target person's name
(e.g., ``\textit{Confirming the annual salary usd for Aiko
Tanaka.}''). \textbf{Multi-hop} questions present only the
person's image and require the model to recall information
without being told the identity (e.g., ``\textit{Based on the
image, what is the person's annual salary usd?}''). Our strict
setting assumes no prior knowledge of the target identity at
query time, which matches the multi-hop format. Furthermore,
Remem~\citep{kwon2026before_remem} notes that single-hop questions
were introduced primarily to induce strong memorization during
vanilla SFT. We therefore report multi-hop in the main text.

\paragraph{Vanilla checkpoint construction.}
For LLaVA-1.5-7B on MLLMU-Bench, we use the publicly released
fine-tuned checkpoint from HuggingFace.\footnote{
\href{https://huggingface.co/yejinkim/mllmu-vanilla}
{MLLMU vanilla checkpoint}} For LLaVA-1.5-7B on
ReMem and Qwen3-VL-8B on MLLMU-Bench, no public vanilla
checkpoint is available, so we fine-tune the base model
in-house following the standard SFT recipe of each benchmark.
We also release some AIM checkpoints to facilitate reproducibility.\footnote{
\href{https://huggingface.co/WonjunLee/AIM_MLLM_Unlearning}
{AIM checkpoints}}

\subsection{Training Details and Hyperparameters}
\label{app:hyperparam}

\paragraph{Hardware.}
All experiments were conducted on the following environment:
\begin{itemize}
\item GPU: NVIDIA A100 80GB PCIe
\item CPU: AMD EPYC 7763 64-Core
\item OS: Ubuntu 22.04.5 LTS, kernel 6.8.0-111-generic, x86\_64
\item CUDA: 12.4 / PyTorch: 2.6.0+cu124
\end{itemize}

\paragraph{Stage 1 (VP learning).}
\begin{itemize}
\item Learning rate $\eta_T$: $1e^{-2}$
\item Batch size: 4
\item Iterations: 10 epochs
\item Loss coefficients: 
$(\lambda_{\mathrm{CE}} : \lambda_{\mathrm{align}} : \lambda_{\mathrm{norm}}) 
= (1 : 10 : 0.1)$ unless stated otherwise.
\item IDK target set $\mathcal{Y}_{\mathrm{idk}}$: Provided in the supplementary material.
\item Use 8 common attributes in every forget set and create questions using GPT-4o-mini.
\end{itemize}

\paragraph{Stage 2 (Fisher-constrained update).}
\begin{itemize}
\item Damping for Fisher inversion: 
$\mathbb{E}[\widetilde{\mathbf{F}}_r]$
\item Effective Learning rate: $1e^{-5}$
\item Iterations: 50 epochs
\item Batch size: 4
\end{itemize}

\paragraph{GPT-based evaluation.}
All GPT-based judgments in this work (same-meaning, 
appropriateness, and other GPT-judge metrics) use 
\texttt{gpt-4o-mini}. The prompts used for evaluation are provided in the supplementary material.

\subsection{Stage 1 Loss-Weight Ablation}
\label{app_sec:stage1_loss_weight}
\begin{table}[t]
\centering
\resizebox{\columnwidth}{!}{%
\begin{tabular}{lccccc}
\toprule
$\lambda_\text{CE}{:}\lambda_\text{align}{:}\lambda_\text{norm}$ & EM$_f$$\downarrow$ & EM$_r$$\uparrow$ & ROUGE$\uparrow$ & Exp$\downarrow$ & EM$_t$$\downarrow$ \\
\midrule
$1{:}1{:}1$ & 76.7 & 97.8 & 98.7 & 68.9 & 33.9 \\
$1{:}10{:}1$ & 75.0 & 95.4 & 97.6 & 68.9 & 63.1 \\
$1{:}1{:}0.1$ & 10.0 & 38.6 & 68.3 & 59.0 & 58.3 \\
$1{:}10{:}0.1$ (baseline) & 38.3 & 78.8 & 89.2 & 63.9 & 35.1 \\
\bottomrule
\end{tabular}%
}
\caption{\textbf{Stage 1 loss-weight ablation (LLaVA-1.5-7B, ReMem, 10\%).} Weights are
$(\lambda_\text{CE}{:}\lambda_\text{align}{:}\lambda_\text{norm})$ in Eq.~(\ref{eq:vp_loss}), IDK trigger fixed.}
\label{app_tab:abl-weights}
\end{table}

Table~\ref{app_tab:abl-weights} ablates the three Stage 1
loss weights $(\lambda_{\mathrm{CE}}:\lambda_{\mathrm{align}}:
\lambda_{\mathrm{norm}})$ in Eq.~(\ref{eq:vp_loss}) on
LLaVA-1.5-7B / ReMem at the 10\% forget split, with the IDK
trigger fixed. The baseline $(1{:}10{:}0.1)$ achieves a balanced
forget/retain trade-off. Increasing $\lambda_{\mathrm{norm}}$
(rows 1--2) over-constrains the visual prompt and weakens
forget reduction. Reducing $\lambda_{\mathrm{align}}$ (row 3)
yields aggressive forgetting but at the cost of severe retain
degradation.

\subsection{LLaVA-1.5-7B at Forget 15\%}
\label{app_sec:llava_15}
\begin{table*}[t]
\centering
\footnotesize
\resizebox{\textwidth}{!}{%
\begin{tabular}{cl|cccccc|ccccc}
\toprule
& & \multicolumn{6}{c|}{MLLMU-Bench} & \multicolumn{5}{c}{ReMem} \\
\cmidrule(lr){3-8} \cmidrule(lr){9-13}
& Method & Cls$_f\downarrow$ & ROUGE$_f\downarrow$ & Cls$_r\uparrow$ & ROUGE$_r\uparrow$ & Cls$_c\uparrow$ & ROUGE$_c\uparrow$ & EM$_f\downarrow$ & EM$_r\uparrow$ & ROUGE$\uparrow$ & Exp$\downarrow$ & EM$_t\downarrow$ \\

\midrule
\multicolumn{13}{c}{\textit{Forget 15\%}} \\
\midrule
& Vanilla       & 42.1 & 56.1 & 38.5 & 55.5 & 52.3 & 24.4 & 100.0 & 100.0 & 100.0 & 68.2 & 100.0 \\
\cmidrule(lr){1-13}
\multirow{4}{*}{\rotatebox[origin=c]{90}{\shortstack{retain\\utilizing}}}
& GA\_Diff      & 39.7 & 47.0 & 36.6 & 52.4 & 51.0 & 23.0 &  0.0 &  94.7 &   32.3 & 35.9 &  0.0 \\
& KL\_Min       & 24.0 & 11.1 & 37.4 & 38.0 & 54.4 & 14.7 &  0.0 &   0.0 &   0.0 & 47.6 &  0.0 \\
& MMUnlearner   & 35.7 & 45.2 & 36.1 & 50.6 & 37.7 & 24.2 &  2.2 &  81.9 &  91.0 & 45.3 &  0.0 \\
& MANU          & 39.2 & 55.5 & 36.4 & 51.7 & 52.0 & 23.9 & 72.2 & 99.0 & 98.3 & 65.9 & 75.4  \\
\cmidrule(lr){1-13}
\multirow{3}{*}{\rotatebox[origin=c]{90}{\shortstack{forget\\only}}}
& GA            &  0.0 &  0.0 &  0.0 &  0.0 &  0.0 &  0.0 &  0.0 &   0.0 &   0.0 & 48.6 &  0.0 \\
& NPO           & 32.5 &  0.9 & 27.5 &  1.0 & 10.8 &  1.1 &  0.0 &   0.0 &   0.0 & 51.6 &  0.0 \\
& \hl{\textbf{Ours}} & \hl{38.9} & \hl{49.0} & \hl{38.2} & \hl{52.0} & \hl{50.4} & \hl{27.8} & \hl{51.1} & \hl{72.5} & \hl{85.6} & \hl{63.2} & \hl{48.0} \\

\bottomrule
\end{tabular}%
}
\caption{\textbf{Forget 15\% results on LLaVA-1.5-7B.}
Continuation of Table~\ref{tab:combined_all} for the 15\% forget
split, with the same baseline groupings (retain-utilizing,
forget-only, ours). Subscripts $f$, $r$, $c$, $t$ denote
forget, retain, celebrity, and test splits. Exp is the
ReMem Exposure metric.}
\label{app_tab:llava_15}
\end{table*}

Table~\ref{app_tab:llava_15} reports the Forget 15\% results
on LLaVA-1.5-7B that were omitted from Table~\ref{tab:combined_all}
for space. AIM continues to exhibit the balanced
forget/retain trade-off observed at 5\% and 10\%, reducing
forget metrics on both MLLMU-Bench and ReMem while keeping
retain performance competitive among methods that achieve
meaningful forget reduction.

\subsection{Qwen3-VL-8B Implementation}
Qwen3-VL adopts a DeepStack architecture in which intermediate vision-encoder layers are injected into corresponding intermediate LLM layers, alongside the standard projection of the final visual features. AIM targets the final encoder feature through $\mathcal{L}_{\mathrm{align}}$ in Stage~1 and the feature-matching loss in Stage~2. Applying these only to the final feature would leave the intermediate visual signals unaltered, allowing identity information to leak through the DeepStack injections. We therefore extend both losses to every intermediate vision feature injected into the LLM: $\mathcal{L}_{\mathrm{align}}$ aligns displacements at each injected layer, and the Stage~2 loss matches the post-update intermediate feature to its prompted target at the same layer.


\subsection{Additional Evaluation Results on Various Models}
\label{app_sec:qwen_15_and_resize}
\begin{table}[t]
\centering
\footnotesize
\resizebox{\columnwidth}{!}{%
\begin{tabular}{l|cccccc}
\toprule
Method & Cls$_f\downarrow$ & R$_f\downarrow$ & Cls$_r\uparrow$ & R$_r\uparrow$ & Cls$_c\uparrow$ & R$_c\uparrow$ \\

\midrule
\multicolumn{7}{c}{\textit{Forget 15\%}} \\
\midrule
Vanilla       & 70.1 & 68.2 & 68.8 & 71.3 & 74.4 & 43.7 \\ \midrule
\textbf{\textit{retain-utilizing}} & & & & & & \\
GA\_Diff      & 65.9 & 58.7 & 66.5 & 67.3 & 73.4 & 42.2 \\
KL\_Min       &  5.6 &  5.4 & 54.9 & 63.6 & 68.5 & 35.4 \\
MMUnlearner & 47.7 & 46.9 & 57.0 & 54.0 & 58.4 & 36.3 \\
MANU & 58.1 & 56.7 & 60.0 & 57.8 & 68.8 & 41.4 \\ \midrule
\textbf{\textit{forget-only}} & & & & & & \\
GA            & 24.0 &  0.2 & 22.6 &  0.3 & 50.8 &  0.1 \\
NPO           &  8.5 &  1.8 &  7.0 &  1.9 & 42.3 &  9.1 \\
\hl{\textbf{Ours}} & \hl{66.1} & \hl{63.6} & \hl{66.5} & \hl{68.9} & \hl{74.5} & \hl{43.7} \\

\midrule
\multicolumn{7}{c}{\textit{Native Resolution Inference on Forget 5\%}} \\
\midrule
Vanilla       & 65.6 & 59.1 & 65.2 & 59.9 & 74.4 & 43.5 \\ \midrule
\textbf{\textit{retain-utilizing}} & & & & & \\
GA\_Diff      & 56.8 & 58.0 & 64.4 & 57.7 & 73.8 & 42.7 \\
KL\_Min       & 59.2 & 53.2 & 61.4 & 50.7 & 74.9 & 39.1 \\
MMUnlearner   & 48.4 & 45.0 & 48.0 & 51.9 & 47.4 & 35.4  \\
\midrule \textbf{\textit{forget-only}} & & & & & & \\
GA            & 60.8 & 63.6 & 63.4 & 59.2 & 74.9 & 42.5 \\
NPO           & 19.2 & 43.4 & 21.3 & 40.7 & 52.1 & 37.0 \\
\hl{\textbf{Ours}} & \hl{58.8} & \hl{55.5} & \hl{65.5} & \hl{59.6} & \hl{74.5} & \hl{43.7} \\

\bottomrule
\end{tabular}%
}
\caption{\textbf{Forget 15\% results and native-resolution
inference on Qwen3-VL-8B.} Top: Forget 15\% results omitted
from Table~\ref{tab:qwen3_mllmu} for space. Bottom: native-resolution
inference at Forget 5\%. Baseline groupings follow Table~\ref{tab:qwen3_mllmu}.}
\label{app_tab:qwen3_mllmu_resize}
\end{table}
\begin{table}[t]
\centering
\footnotesize
\resizebox{\columnwidth}{!}{%
\begin{tabular}{l|cc@{\hspace{1.2em}}cc}
\toprule

& \multicolumn{2}{c}{\textit{LLaVA-1.5-13B}}
& \multicolumn{2}{c}{\textit{InternVL3-8B}} \\
\cmidrule(lr){2-3} \cmidrule(lr){4-5}
Method & ROUGE$_f\downarrow$ & ROUGE$_r\uparrow$
& ROUGE$_f\downarrow$ & ROUGE$_r\uparrow$ \\
\midrule

Vanilla & 62.2 & 59.9 & 67.9 & 67.6 \\ \midrule

\textbf{\textit{retain-utilizing}} & & & & \\
KL\_Min     & 46.6 & 55.1 & 66.2 & 68.4 \\
MMUnlearner & OOM  & OOM  & 52.8 & 59.9 \\ \midrule

\textbf{\textit{forget-only}} & & & & \\
GA  & 0.1 & 0.1 & 0.0 & 0.0 \\
NPO & 4.6 & 4.4 & 7.9 & 8.4 \\
\hl{\textbf{Ours}}
& \hl{53.7} & \hl{57.4}
& \hl{55.2} & \hl{62.1} \\

\bottomrule
\end{tabular}%
}
\caption{\textbf{Additional backbone results on MLLMU-Bench at the 10\% forget split.}
AIM achieves a competitive forget--retain trade-off on both LLaVA-1.5-13B and InternVL3-8B.}
\label{app_tab:additional_backbones}
\end{table}

Table~\ref{app_tab:qwen3_mllmu_resize} reports two
supplementary experiments on Qwen3-VL-8B. The top panel
contains Forget 15\% results omitted from
Table~\ref{tab:qwen3_mllmu} for space and the bottom panel evaluates dynamic-resolution capability.
Unlike LLaVA, the Qwen series natively accepts
variable input sizes, but our main experiments resize all
images to $448\times448$. Therefore, we re-evaluate the unlearned model at the original image sizes (without resizing) on the 5\% forget split. While absolute scores drop slightly under native-resolution inputs, the forget/retain trade-off matches the fixed-resolution
setting, suggesting that AIM largely preserves unlearning ability to process variable-sized inputs.

We also report results for two additional models, LLaVA-1.5-13B and InternVL3-8B, in Table~\ref{app_tab:additional_backbones}. Across both backbones, AIM maintains a competitive forget--retain trade-off, extending the trend observed in the main experiments to larger and alternative architectures.

\subsection{General VQA Performance after Unlearning}
\label{app:vqav2_utility}
\begin{table}[t]
    \centering
    \small
    \setlength{\tabcolsep}{3.5pt}
    \resizebox{\columnwidth}{!}{%
    \begin{tabular}{lcccc}
        \toprule
        \multirow{2}{*}{\textbf{Method}}
        & \multicolumn{2}{c}{\textbf{LLaVA-1.5-7B}}
        & \multicolumn{2}{c}{\textbf{Qwen3-VL-8B}} \\
        \cmidrule(lr){2-3}
        \cmidrule(lr){4-5}
        & \textbf{5\% Forget}
        & \textbf{10\% Forget}
        & \textbf{5\% Forget}
        & \textbf{10\% Forget} \\
        \midrule
        Vanilla
        & 67.6 & 67.6
        & 74.1 & 74.1 \\ \midrule
        MMUnlearner
        & 30.1 & 45.4
        & 49.2 & 14.3 \\
        GA
        & 0.0 & 0.1
        & 75.2 & 73.9 \\
        NPO
        & 59.2 & 0.0
        & 75.5 & 62.2 \\
        \hl{\textbf{Ours}}
        & \hl{65.9} & \hl{60.9}
        & \hl{73.4} & \hl{73.8} \\
        \bottomrule
    \end{tabular}%
    }
    \caption{\textbf{VQAv2-10K accuracy after unlearning.}
    Accuracy is evaluated on 10K randomly sampled examples from the
    VQAv2 validation set.}
    \label{app_tab:vqav2_utility}
\end{table}
Beyond visual perception on the forget-identity images, we further evaluate general visual utility on 10K randomly sampled examples from the VQAv2 validation set~\citep{goyal2017making}.
As shown in Table~\ref{app_tab:vqav2_utility}, AIM largely preserves general VQA performance after unlearning. Overall, these results indicate that AIM generally preserves visual utility beyond the identity images.

\subsection{Continual Unlearning Setup}
\label{app:continual_detail}

The continual unlearning experiment 
(Tab.~\ref{tab:continual_mh}) uses two sequential forget transitions:
$5 \to 10$ and 
$10 \to 15$.

\paragraph{5\% $\to$ 10\%}
Forget 10 is a strict superset of Forget 5 as 
defined by the standard ReMem splits. We continually
unlearn the additional identities (the 5\% complement) starting 
from the forget 5 checkpoint.

\paragraph{10\% $\to$ 15\%}
The standard benchmark splits at 10\% and 15\% are not nested. 
We construct forget 15 ourselves by extending forget 10 with additional held-out identities from the retain pool, preserving the identity-level structure of the 
benchmark. This setup deliberately matches the real-world continual scenario in which new deletion requests need not align with predefined dataset splits.


\section{AI Assistant Usage Statement}
We used an AI assistant for language polishing and proofreading of the manuscript. All research ideas, experimental design, implementation, analysis, and conclusions are the work of the authors.

\begin{table}[t]
\centering
\small
\resizebox{\columnwidth}{!}{
\begin{tabular}{p{0.34\columnwidth}p{0.22\columnwidth}p{0.30\columnwidth}}
\toprule
\textbf{Asset} & \textbf{Type} & \textbf{License} \\
\midrule

MLLMU-Bench & Dataset & Not specified \\
ReMem & Dataset & Not specified \\

\midrule

LLaVA-1.5-7B & Base model & Llama 2 Community License \\
Qwen3-VL-8B-Instruct & Base model & Apache 2.0 \\

\midrule

MMUnlearner & Baseline & Not specified \\
MANU & Baseline & Not specified \\

\bottomrule
\end{tabular}
}
\caption{Licenses of datasets, base models, and baseline methods used in this work.}
\label{tab:licenses}
\end{table}

\section{License of Datasets and Models}
We summarize the licenses of all datasets, pretrained models, and baseline implementations used in this work in Tab.~\ref{tab:licenses}. All assets are used in accordance with their respective licenses.

\end{document}